\documentclass[accepted]{uai2026} 
                        
\usepackage[american]{babel}

\usepackage{natbib} 
\usepackage{mathtools} 
\usepackage{booktabs} 
\usepackage{tikz} 

\usepackage{amsmath}
\usepackage{amssymb}
\usepackage{amsthm}
\usepackage{graphicx}
\usepackage{microtype}
\usepackage{graphicx}
\usepackage{booktabs} 
\usepackage{hyperref}
\usepackage{algorithm}
\usepackage{algorithmic}
\usepackage{placeins}
\usepackage{dblfloatfix}

\usepackage{colortbl}
\usepackage{xcolor}

\definecolor{improvGreen}{RGB}{0,150,0} 
\definecolor{improvRed}{RGB}{180,0,0}   

\definecolor{highrank}{RGB}{198,239,206}   
\definecolor{lowrank}{RGB}{255,199,206}    

\usepackage{listings}
\definecolor{magentaPink}{RGB}{225,30,130}
\definecolor{darkgreen}{RGB}{0,130,0}

\lstdefinestyle{pseudopy}{
  language=Python,
  basicstyle=\ttfamily\footnotesize,
  keywordstyle=\color{magentaPink},
  commentstyle=\color{darkgreen},
  stringstyle=\color{teal},
  showstringspaces=false,
  frame=none,
  tabsize=2,
  breaklines=true,
  columns=fullflexible
}

\numberwithin{equation}{section}
\theoremstyle{plain}
\newtheorem{theorem}{Theorem}[section]

\newtheorem{lemma}[theorem]{Lemma}
\newtheorem{corollary}[theorem]{Corollary}
\theoremstyle{definition}
\newtheorem{definition}[theorem]{Definition}

\theoremstyle{remark}

\usepackage[textsize=tiny]{todonotes}

\usepackage[autostyle, english=american]{csquotes}
\MakeOuterQuote{"}

\usepackage{booktabs}
\usepackage{multirow}
\usepackage{newfloat}
\usepackage{listings}
\usepackage{dsfont}
\usepackage{arydshln}
\usepackage{caption}
\usepackage{subcaption}
\usepackage{lipsum}
\usepackage{bbm}
\usepackage{float}
\usepackage{centernot}

\usepackage{bm}
\usepackage{adjustbox}
\usepackage{makecell}

\DeclarePairedDelimiter\autobracket{(}{)}
\newcommand{\br}[1]{\autobracket*{#1}}

\newcommand{\bx}{\bm{x}}
\newcommand{\bth}{\bm{\theta}}

\newcommand{\independent}{\perp\mkern-9.5mu\perp}
\newcommand{\notindependent}{\centernot{\independent}}

\renewcommand{\Pr}{\mathrm{Pr}}

\usepackage{dcolumn}
\newcolumntype{d}[1]{D{.}{.}{#1}}
\DeclareUnicodeCharacter{2212}{-}

\def\ie{{\em i.e.},\ }
\def\eg{{\em e.g.},\ }
\def\SurvP#1#2{\widehat{S}_E(\,#1 \mid #2\,)}

\def\dd{{\rm d}}

\usepackage[inline]{trackchanges}
\addeditor{RG} 
\addeditor{CL}
\addeditor{SQ}  

\title{Overcoming Dependent Censoring in the Evaluation of Survival Models}

\author[1,2]{\href{mailto:clillelund@ualberta.ca?Subject=Your UAI 2026 paper}{Christian~Marius~Lillelund}{}}
\author[1,3,4]{Shi-ang~Qi}
\author[1,3]{Russell~Greiner}

\affil[1]{%
    Department of Computing Science\\
    University of Alberta\\
    Edmonton, Canada
}
\affil[2]{%
    Department of Electrical and Computer Engineering\\
    Aarhus University\\
    Aarhus, Denmark
}
\affil[3]{%
    Alberta Machine Intelligence Institute\\
    Edmonton, Canada
}
\affil[4]{%
    Vector Institute\\
    Toronto, Canada
}
  
\begin{document}
\maketitle

\begin{abstract}
Dependent censoring occurs when the event time and censoring time are not conditionally independent given the observed covariates. This complicates survival model evaluation because widely used metrics, such as the Brier score, typically handle right-censoring using inverse probability of censoring weighting (IPCW). Unfortunately, IPCW is valid only when the estimated censoring distribution is independent of the event time. We propose a dependent Brier score based on an Archimedean copula and the Copula-Graphic estimator, and establish consistency and asymptotic normality of its margin-time estimator. To evaluate the metric, we introduce a semi-synthetic framework that creates realistic dependent censoring while preserving the original covariate structure and known event times. Across 12 datasets, the proposed metric reduces estimation error by 12-16\% on average relative to IPCW. Source code is available at \url{https://github.com/thecml/DependentEVAL}.
\end{abstract}

\section{Introduction}
\label{sec:introduction}

Survival analysis differs from standard regression because outcomes may be \emph{censored}: for some individuals, we only know that the event time exceeds a certain specified value. A fundamental assumption underlying most methods and evaluation metrics is \emph{independent censoring}, meaning that finding someone censored at time $c$ does not convey information about their true survival time $e$, beyond $e>c$~\citep{kleinbaum2012survival}. Censoring is considered \emph{dependent} whenever the censoring time $C$ and the event time $E$ are not conditionally independent given the observed covariates $X$ -- \ie $E \notindependent C \mid X$ (see Appendix \ref{app:censoring_assumptions} for details). While dependent censoring has been relatively well studied in the context of \emph{model estimation}, with several methods proposed to address it~\citep{Foomani2023,zhang2024deep,liu2025hacsurv}, its impact on \emph{model evaluation} remains largely unaddressed. In particular, the integrated Brier score (IBS)~\citep{graf1999assessment} is typically implemented via inverse probability of censoring weighting (IPCW) based on the Kaplan-Meier (KM)~\citep{kaplan_nonparametric_1958} estimator or a Cox proportional hazards (CoxPH) model~\citep{cox_regression_1972}. These approaches assume independence between event and censoring times. When this assumption is violated, the IPCW weights are misspecified, potentially biasing performance estimates and distorting comparisons between survival models.

\begin{figure*}[!htbp]
  \centering
  \begin{subfigure}[b]{0.3\textwidth}
    \centering
    \includegraphics[width=\textwidth, trim=15 15 15 10, clip]{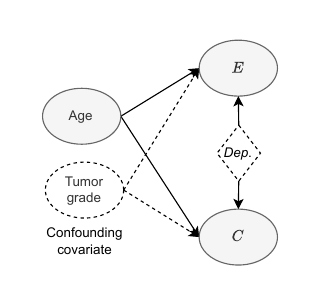}
    \caption{}
    \label{fig:dependence1}
  \end{subfigure}
  \begin{subfigure}[b]{0.36\textwidth}
    \centering
    \includegraphics[width=\textwidth, trim=15 20 20 10, clip]{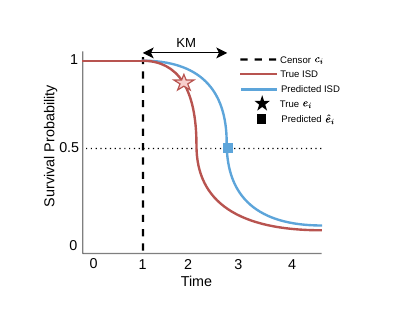}
    \caption{}
    \label{fig:dependence2}
  \end{subfigure}
  \begin{subfigure}[b]{0.33\textwidth}
    \centering
    \includegraphics[width=\textwidth]{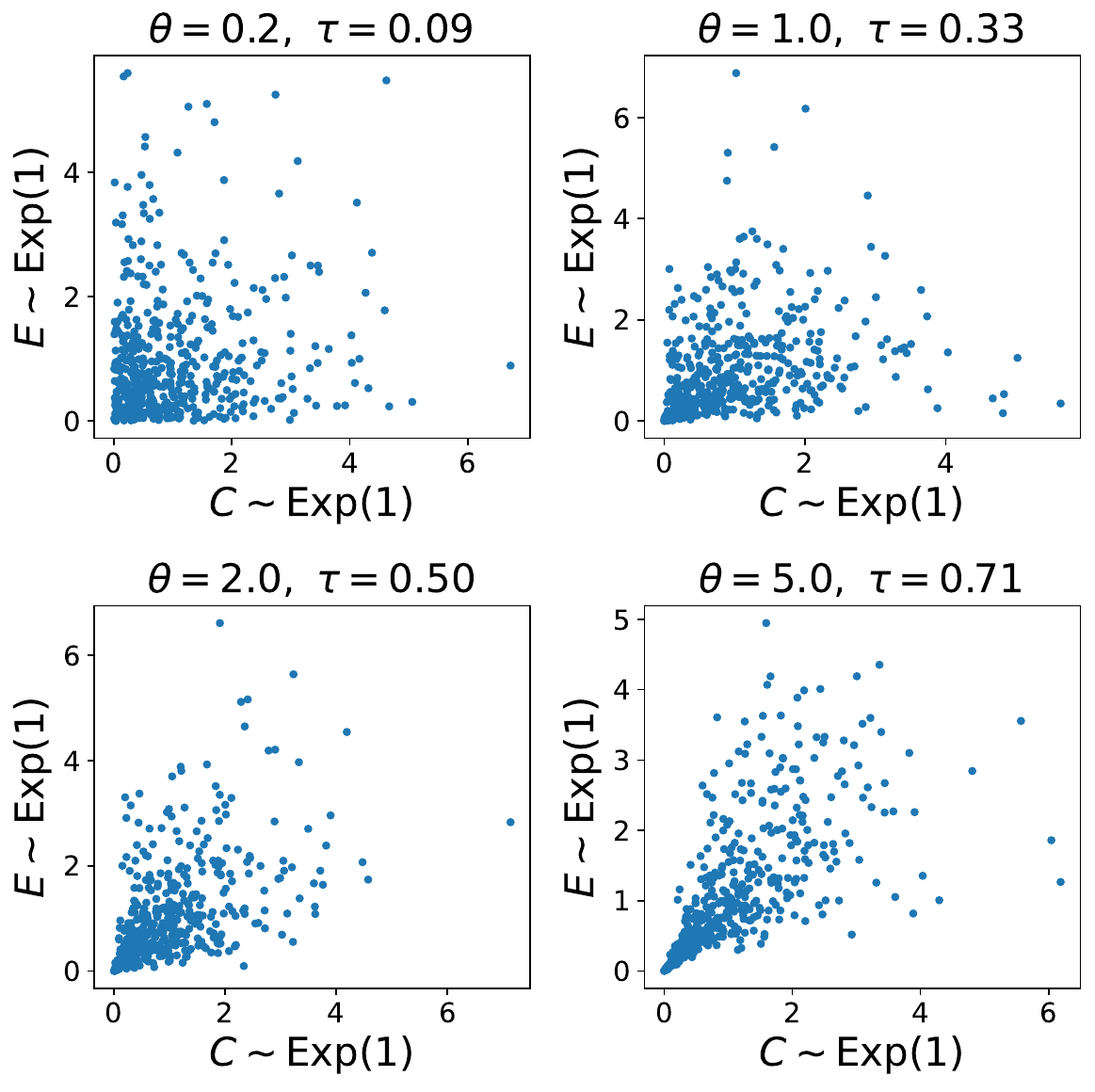}
    \caption{}
    \label{fig:dependence3}
  \end{subfigure}
  \caption{(a)~Residual dependence between event time $E$ (cancer relapse) and censoring time $C$ (dropout) denoted by the diamond "Dep.", due to an unobserved confounding covariate -- here "Tumor grade".
  (b)~Evaluation under dependent censoring, where a patient $i$ is censored at time \(c_i\) just prior to the event, \(e_i\), due to reasons related to the event (but unobserved). This induces bias in the predicted individual survival distribution and the predicted event time $\widehat{e_i}$ produced by the KM estimator, because the censored patient is assumed to be similar to those still at risk. Note that the true and KM-estimated survival curves coincide on the interval \( [0, c_i] \) -- here $c_i=1$ -- but diverge beyond due to dependent censoring.
  (c)~Scatter plots of $(e_i, c_i)$, $i=1,\dots,500$, simulated from a Clayton copula~\citep{clayton1978model} with parameter $\theta$ (reported via the corresponding Kendall’s $\tau$) and standard exponential marginals. Samples are generated by drawing $(u_{1i},u_{2i}) \sim C_\theta$ and applying the inverse exponential CDF. The dependence structure is induced by the Archimedean generator $\varphi_\theta$, where $ C_\theta(u_1,u_2) = \varphi_\theta^{-1}\!\bigl(\varphi_\theta(u_1)+\varphi_\theta(u_2)\bigr).$ This is the same generator function that appears in the CG estimator in Definition~\ref{def:cg_estimator}.}
  \label{fig:dependence}
\end{figure*}

Figure~\ref{fig:dependence1} illustrates how the independent censoring assumption fails when a confounding covariate -- such as "Tumor grade" -- is not observed. Because tumor grade affects both relapse ($E$) and dropout ($C$), advanced tumor grades lead to shorter event and censoring times, inducing dependence. Standard KM-based evaluation relies on independent censoring, assuming that censored individuals are like those remaining at risk. As shown in Figure~\ref{fig:dependence2}, when a patient $i$ is censored before the event for reasons related to the event, the KM estimator may overestimate survival beyond \(c_i\). Because this estimated survival distribution is used to construct the IPCW weights, the distortion produces misspecified weights and, potentially, biased performance estimates.

\textbf{Contribution.}
We introduce a dependent Brier score and its integrated version, IBS-Dep, for evaluating survival models under dependent censoring. Our approach models the joint distribution of event and censoring times and uses the Copula-Graphic (CG) estimator to impute (marginal) event times for censored instances. Our theory establishes consistency and asymptotic normality of the CG-based margin-time estimator; consequently, IBS-Dep is consistent for the corresponding margin-time surrogate Brier risk. We evaluate how well this surrogate approximates the oracle IBS based on uncensored event times in a semi-synthetic framework, and find that IBS-Dep reduces estimation error compared to IBS using IPCW. To our knowledge, this is the first metric for survival analysis that supports dependent censoring.

\section{Background and related work}
\label{sec:background_and_related_work}

\subsection{Survival data}

We consider a survival dataset $\mathcal{D} = \{(\bm{x}_i, t_i, \delta_i)\}_{i=1}^N$, where for each instance $i$, $\bm{x}_i \in \mathbb{R}^d$ denotes the covariates, $t_i \in \mathbb{R}_+$ denotes the observed time (either event or censoring), and $\delta_i \in \{0,1\}$ denotes the event indicator. Each instance has latent event and censoring times, $e_i$ and $c_i$, with $t_i = \min(e_i, c_i)$ and $\delta_i = \mathbf{1}\{e_i \le c_i\}$. At the population-level, let $E$ and $C$ denote non-negative random variables for the true event and censoring times, with cumulative distribution functions $F_E(t)=\Pr(E \le t)$ and $F_C(t)=\Pr(C \le t)$ and survival functions $S_E(t)=1-F_E(t)$ and $S_C(t)=1-F_C(t)$. A common nonparametric estimator of $S_E(t)$ is the KM estimator.

\begin{definition}
Let $u_1<\cdots<u_J$ denote the distinct observed event times, and let
$t_{\max}=\max_{1\leq i\leq N} t_i$. For each event time $u_j$, define
$d_j = \sum_{i=1}^{N} \mathbbm{1}[t_i=u_j,\delta_i=1]$ as the number of events at time $u_j$, and $ n_j = \sum_{i=1}^{N} \mathbbm{1}[t_i\geq u_j] $
as the number of instances at risk immediately before $u_j$. The
Kaplan-Meier (KM) estimator is
\begin{equation}
\label{def:km_estimator}
\widehat{S}_E(t) \
= \
\prod_{u_j\leq t}
\left(
1-\frac{d_j}{n_j}
\right),
\qquad
0 \ \leq  \ t \ \leq \ t_{\max}.
\end{equation}
The empty product is defined as one, and $\widehat{S}_E(t)$ is undefined
for $t>t_{\max}$. See Appendix~\ref{app:km_estimator} for the derivation.
\end{definition}

\subsection{Copula-based estimation}
\label{sec:depCensor}

\citet{Zheng1995} generalized the KM estimator to the CG estimator, which relaxes its independent censoring assumption by modeling dependence through a copula. A copula links two marginal distributions into a joint distribution and can thereby capture dependence between $E$ and $C$. \citet{Rivest2001} derived the first explicit form of the CG estimator under an \emph{assumed copula} from the Archimedean family (\eg the Clayton~\citep{clayton1978model} and Frank~\citep{frank1979simultaneous} copulas), which specifies dependence through a single parameter $\theta$ via a generator function $\varphi_\theta$~\citep{Emura2018} (see details in Appendix \ref{app:copulas_for_survival}). Figure~\ref{fig:dependence3} shows sampling from a Clayton copula, which can model \emph{lower-tail dependence} -- that is, a tendency for early events (small values of $E$) to occur alongside early censoring (small values of $C$). Now, assuming an Archimedean copula, we can estimate $\widehat{S}_E(t)$ using the CG estimator.

\begin{definition}
Let $u_1<\cdots<u_J$ denote the distinct observed event times, and let $t_{\max}=\max_{1\leq i\leq N} t_i$. The Copula-Graphic (CG) estimator is
\begin{equation}
\label{def:cg_estimator}
\widehat{S}_E(t) \
= \
\varphi_\theta^{-1}
\left(
\sum_{u_j\leq t}
\left[
\varphi_\theta\!\left(\frac{n_j-d_j}{N}\right)
-
\varphi_\theta\!\left(\frac{n_j}{N}\right)
\right]
\right).
\end{equation}
For $0\leq t\leq t_{\max}$, where $\varphi_\theta$ is an Archimedean copula generator, $d_j=\sum_{i=1}^{N}\mathbbm{1}[t_i=u_j,\delta_i=1]$ is the number of events at $u_j$, and $n_j=\sum_{i=1}^{N}\mathbbm{1}[t_i\geq u_j]$ is the number at risk immediately before $u_j$. The empty sum is zero, and $\widehat{S}_E(t)$ is undefined for $t>t_{\max}$~\citep{Emura2018}. For independent copula $\varphi_{\mathrm{ind}}(u)=-\log u$, the CG estimator reduces to the KM estimator. See Appendix~\ref{app:cg_estimator} for the derivation.
\end{definition}

\subsection{IPCW-weighted Brier score}
\label{sec:Eval-bias}

The Brier score (BS) measures the squared error between the predicted survival probability and the observed binary outcome at a time $t^*$~\citep{Brier1950}, which for uncensored instances, is:
\begin{equation}
\label{eq:bs_uncensored}
\mathrm{BS}\bigl(
\mathcal{D};\widehat{S}_E(t^*\mid\cdot)
\bigr)
=
\frac{1}{N}\sum_i
\bigl(
\mathbbm{1}[t_i>t^*]
-
\SurvP{t^*}{\bm{x}_i}
\bigr)^2.
\end{equation}
To incorporate censoring, \citet{graf1999assessment} used inverse probability of censoring weights (IPCW), where $\widehat{S}_C(t)$ is typically estimated using the marginal KM estimator in standard implementations~\citep{qi_survivaleval_2024}\footnote{More generally, it may use a conditional estimator $\widehat{S}_C(t \mid X)$.}:
\begin{align}
w_i\!\left(t^* \right) \ = \ \frac{\delta_i \, \mathbbm{1}[t_i \leq t^*]}{\widehat{S}_C(t_i)} + \frac{\mathbbm{1}[t_i > t^*]}{\widehat{S}_C(t^*)},
\end{align}
where the first term applies to events before $t^*$ and the second to those surviving beyond. Individuals censored before $t^*$ receive weight zero and therefore do not contribute to the sum. Then, the BS-IPCW metric can be expressed as:
\begin{equation}
\label{eq:bs_ipcw}
\begin{aligned}
& \mathrm{BS}_{\mathrm{IPCW}}\bigl(
\mathcal{D}; \widehat{S}_E(t^* \mid \cdot)
\bigr) \\
&\ = \ \frac{1}{N} \sum_{i=1}^N w_i(t^*)  \left( \mathbbm{1}[t_i > t^*]
- \SurvP{t^*}{\bm{x}_i} \right)^2 .
\end{aligned}
\end{equation}
Often, practitioners want to summarize Equation~\ref{eq:bs_ipcw} over a time interval. Therefore, \citet{graf1999assessment} proposed the integrated Brier score (IBS), defined as the uniform time average of the Brier score over a prespecified interval $[t_1,t_2]$:
\begin{equation}
\label{eq:ibs_ipcw}
\mathrm{IBS}\bigl(
\mathcal{D}; \widehat{S}_E, t_1, t_2
\bigr)
=
\frac{1}{t_2-t_1}
\int_{t_1}^{t_2}
\mathrm{BS}_{\mathrm{IPCW}}\bigl(
\widehat{S}_E(\tau\mid\cdot)
\bigr)
\,\dd\tau.
\end{equation}
A common choice is $[t_1,t_2]=[0,t_{\max}]$, where $t_{\max}$ is the maximum observed time in the dataset.

The IPCW-weighted Brier score depends on estimating the censoring distribution. When $\widehat{S}_C(t)$ is obtained via the KM, its consistency relies on the independent censoring assumption. Under dependent censoring, KM can be biased and inconsistent~\citep{Emura2018,campigotto2014impact}. Conditional IPCW estimators~\citep{gerds2006consistent} attempt to address this by estimating $\widehat{S}_C(t \mid X)$, but two limitations remain: (1)~\emph{model misspecification}, which may give unreliable weights, and (2)~\emph{residual dependence}, where unobserved factors induce dependence not captured by the conditional model.

\section{The dependent Brier score}
\label{sec:proposed_metrics}

\subsection{Definition}
\label{sec:definition}

We propose a Brier score that relaxes independent censoring by modeling dependence via an Archimedean copula. Our method consists of two parts: (1)~the CG estimator to impute \emph{margin time} estimates for censored instances, and (2)~an \emph{uncertainty weighting} mechanism.

The margin time is the conditional expectation of the event time given that the event time is greater than the censoring time. Given that an instance is censored at time $t_i$, we can calculate its margin time using~\citep{haider_effective_2020}:
\begin{equation}
e^{\mathrm{margin}}\!\left(S_E, t_i\right) \ = \ \mathbb{E}[E \mid E > t_i] \ = \ t_i + \frac{\int_{t_i}^{\infty} S_E(t)\,dt}{S_E(t_i)}.
\end{equation}

Because $S_E$ is unknown, we estimate the margin time using $e^{\mathrm{margin}}\!\left(\widehat{S}_E^{\mathrm{CG}}, t_i\right)$,\footnote{For brevity, we write $e^{\mathrm{margin}}(t_i)$ when the survival estimator is fixed.}
where $\widehat{S}_E^{\mathrm{CG}}$ is obtained from the CG estimator (Definition \ref{def:cg_estimator}) under an Archimedean copula with parameter $\theta$. The resulting margin time serves as an imputed event time for each censored instance. This construction requires specification of a copula family and its parameter and accounts for marginal dependence between $E$ and $C$ through their joint copula model.

To account for uncertainty in these imputed event times, we adaptively down-weight censored instances according to how early censoring occurs. For an instance censored at $t_i$, we use its estimated progress through the event-time distribution, $\widehat{F}_E^{\mathrm{CG}}(t_i) = 1-\widehat{S}_E^{\mathrm{CG}}(t_i)$, as an uncertainty weight. We therefore define $w_i = 1$ for uncensored instances and $w_i = 1-\widehat{S}_E^{\mathrm{CG}}(t_i)$ for censored instances. Early censoring yields a smaller $w_i$, whereas later censoring generally yields a larger weight. This leads to the proposed Brier score, which is free from IPCW and targets a margin-time surrogate of the Brier risk:
\begin{equation}
\label{eq:bs_dependent}
\begin{aligned}
&\mathrm{BS}_{\mathrm{Dep}}\!\left(
\mathcal{D};
\widehat{S}_E(t^*\mid\cdot)
\right) \\
&\ = \
\frac{1}{\sum_{i=1}^N w_i}
\sum_{i=1}^N w_i
\Bigl(
\mathbbm{1}[\tilde{e}_i>t^*]
-
\widehat{S}_E(t^*\mid\bm{x}_i)
\Bigr)^2,
\\
\end{aligned}
\end{equation}
where
\begin{equation}
\begin{aligned}
\tilde{e}_i
\ = \
\begin{cases}
e_i, & \delta_i=1, \\
e^{\mathrm{margin}}\!\left(\widehat{S}_E^{\mathrm{CG}}, t_i\right), & \delta_i=0. \notag
\end{cases}
\end{aligned}
\end{equation}

Similarly, we define the integrated dependent Brier score by averaging Equation~\ref{eq:bs_dependent} over $[t_1,t_2]$:
\begin{equation}
\mathrm{IBS}_{\mathrm{Dep}}\bigl(
\mathcal{D}; \widehat{S}_E, t_1, t_2
\bigr)
=
\frac{1}{t_2-t_1}
\int_{t_1}^{t_2}
\mathrm{BS}_{\mathrm{Dep}}\bigl(
\widehat{S}_E(\tau\mid\cdot)
\bigr)
\,\dd\tau.
\end{equation}

Section~\ref{sec:theoretical_analysis} establishes consistency of the CG-based margin-time estimator for its copula-induced target, which implies consistency of IBS-Dep for the associated surrogate Brier risk.

\subsection{Theoretical analysis}
\label{sec:theoretical_analysis}

We now develop a concise large-sample theory for the CG-based margin-time functional under dependent censoring~\citep{Rivest2001}. The theory concerns the imputed margin times used by IBS-Dep, rather than the oracle Brier score based on unobserved event times. For ease of understanding, we sketch the main ideas and implications here, and defer the formal definitions, regularity conditions, and detailed proofs to Appendix~\ref{app:theoretical_properties}.

First, we show that the margin time admits a direct probabilistic interpretation: it equals the conditional mean remaining event time beyond a landmark,
\begin{align*}
    e^{\mathrm{margin}}(c) \ = \  \mathbb{E}[E\mid E > c],
\end{align*}
provided $S_E(c)>0$ and the corresponding tail integral is finite (see Appendix~\ref{app:conditional_exp}). This justifies margin time as a clinically interpretable summary of the post-$c$ prognosis.

Second, we establish \emph{consistency} of the plug-in estimator $\widehat e^{\mathrm{margin}}(c)$ for a large-sample target $e^{*,\mathrm{margin}}(c)$ induced by the (possibly misspecified) Archimedean copula model considered by \citet{Rivest2001}. Under uniform consistency of $\widehat S_E$ for $S_E^*$ on a finite horizon $[0,t_{\max})$ and mild regularity ($S_E^*(c)>0$ and negligible tail contribution beyond $t_{\max}$), a deterministic perturbation bound shows that small uniform errors in $\widehat S_E$ translate into small error in $\widehat e^{\mathrm{margin}}(c)$. When the copula model is correctly specified, $S_E^*=S_E$ and hence $e^{*,\mathrm{margin}}(c)=e^{\mathrm{margin}}(c)$ (see details in Appendix~\ref{app:consistency}).

Finally, we derive a \emph{first-order (delta-method) expansion} and an asymptotic normal limit. Writing the margin time as a functional $\mathcal{T}_c(S)$, we obtain an exact linearization
\begin{align*}
    \mathcal{T}_c(S+h)-\mathcal{T}_c(S) \ = \ \dot{\mathcal{T}}_{c;S}(h) \;+\; R_{c;S}(h),
\end{align*}
with an explicit remainder $R_{c;S}(h)$ that is second order in the perturbation (see Appendix~\ref{app:delta_method_full}). Combining this expansion with the functional central limit theorem (CLT) of \citet[Theorem~2]{Rivest2001} for $\sqrt{n}(\widehat S_E-S_E^*)$ yields the asymptotic linear representation (converges in distribution)
\begin{align*}
    \sqrt{n} & \big(\widehat e^{\mathrm{margin}}(c)-e^{*,\mathrm{margin}}(c)\big)
    \ \Longrightarrow\\
    &\frac{1}{S_E^*(c)}\int_c^{t_{\max}}\mathbb{G}_E(u)\,du
    \;-\;
    \frac{\int_c^{t_{\max}} S_E^*(u)\,du}{(S_E^*(c))^2}\,\mathbb{G}_E(c),
\end{align*}
where $\mathbb{G}_E(c)$ stands for the Gaussian process. If the right-hand side is non-degenerate, this implies an $N^{-1/2}$ convergence rate and enables principled uncertainty quantification for margin time via the induced influence form.

\section{Synthetic experiments}

\subsection{Experimental setup}
\label{sec:synthetic_setup}

To validate our dependent Brier score, we simulate synthetic survival data with known ground-truth event times. Covariates $x_i \sim \mathrm{Unif}([0,1]^d)$ ($d=10$) are fixed per seed. Event and censoring times follow covariate-dependent Weibull proportional hazards models with linear risk functions $g(\bm{x})=\bm{x}^\top\bm{\beta}$ and fixed shape/scale parameters (see Appendix~\ref{app:synthetic_experiments}). Dependence is parameterized by Kendall's $\tau$: when $\tau=0$, event and censoring times are independent; for $\tau>0$, dependence is introduced via a copula, with family and parameter assumed known to the dependent metric. For each of 10 random seeds, we generate $N=10{,}000$ and use fixed training (70\%) and test (30\%) splits. The event times and fitted CoxPH survival model are held fixed across censoring settings, so changes in evaluation error reflect the censoring mechanism rather than changes in the prediction model. Oracle IBS is computed on the test set using the fully observed event times prior to censoring.

\subsection{Results and takeaways}
\label{sec:synthetic_results}

\begin{figure*}[!ht]
\centering
\subfloat[Increasing dependence (Censoring = 50\%).]{
    \includegraphics[width=0.49\linewidth]{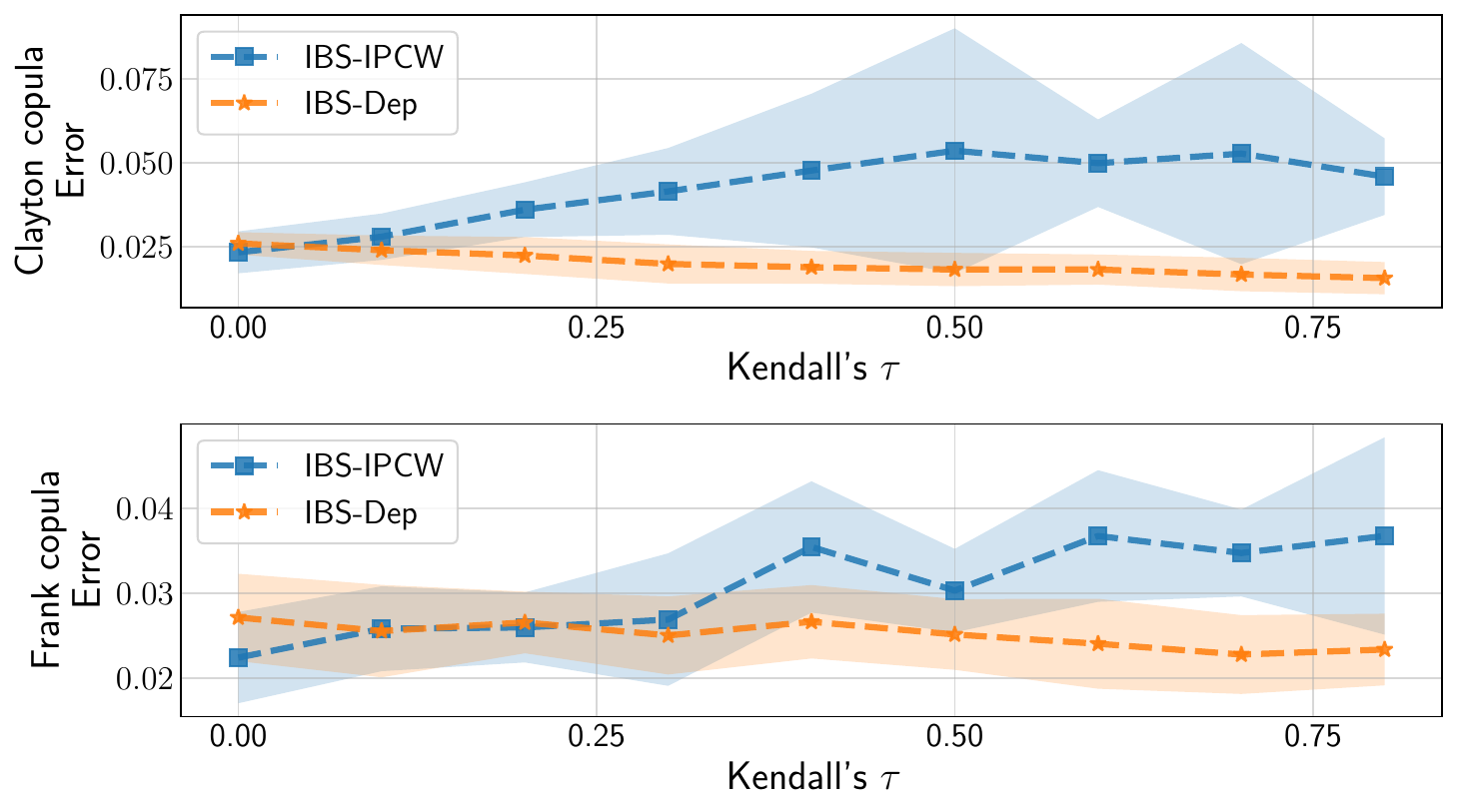}
}
\subfloat[Increasing censoring (Kendall's $\tau = 0.5$).]{
    \includegraphics[width=0.49\linewidth]{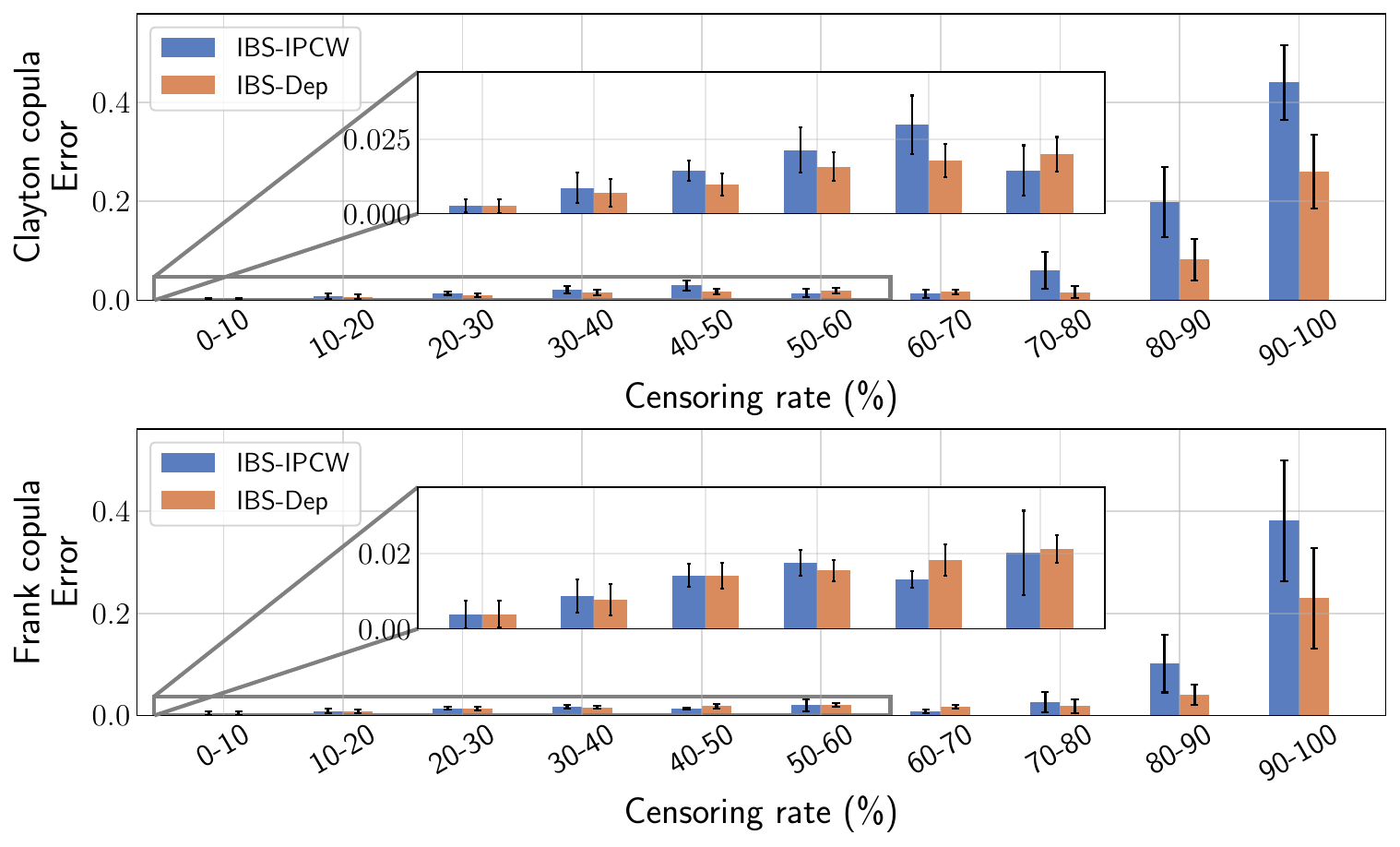}
}
\caption{Mean ($\pm$ SD) absolute estimation error of IBS-IPCW and the proposed IBS-Dep under a \textit{known} copula, averaged over 10 experiments. (a) Error as a function of the dependence between $E$ and $C$ with censoring held constant. (b) Error as a function of the censoring rate with Kendall's $\tau$ held constant.}
\label{fig:synthetic_estimation_error}
\end{figure*}

\textbf{Increasing dependence.} Figure~\ref{fig:synthetic_estimation_error} (a) shows the estimation error as a function of Kendall's $\tau$ at fixed censoring ($\approx 50\%$). IBS-IPCW becomes increasingly unstable as $\tau$ grows, while
our IBS-Dep remains comparatively stable and achieves lower (or comparable) error. This reflects the breakdown of independent censoring: as dependence increases, IPCW becomes biased and exhibits higher variance because marginal inverse weights no longer correct the induced selection. The effect is stronger under the Clayton copula due to its lower-tail dependence, which increases the probability that early (high-risk) events coincide with early censoring and thereby more severely distorts the observable risk set. As a result, the empirical IBS-IPCW deviates increasingly from the oracle IBS computed from the fully observed data, whereas IBS-Dep tracks the oracle much more closely.

\textbf{Increasing censoring.} Figure~\ref{fig:synthetic_estimation_error} (b) shows error as a function of the censoring rate with $\tau=0.5$. Both metrics deteriorate under heavy censoring, but IBS-IPCW degrades much more rapidly, particularly beyond $\sim 80\%$ censoring. As censoring increases, IPCW relies on fewer uncensored observations with larger inverse weights, inflating variance and bias under dependence. In contrast, IBS-Dep shows slower error growth, since it adjusts contributions using the joint $(E,C)$ structure rather than relying solely on extreme reweighting.

\begin{figure}[!ht]
\centering
\includegraphics[width=1\linewidth]{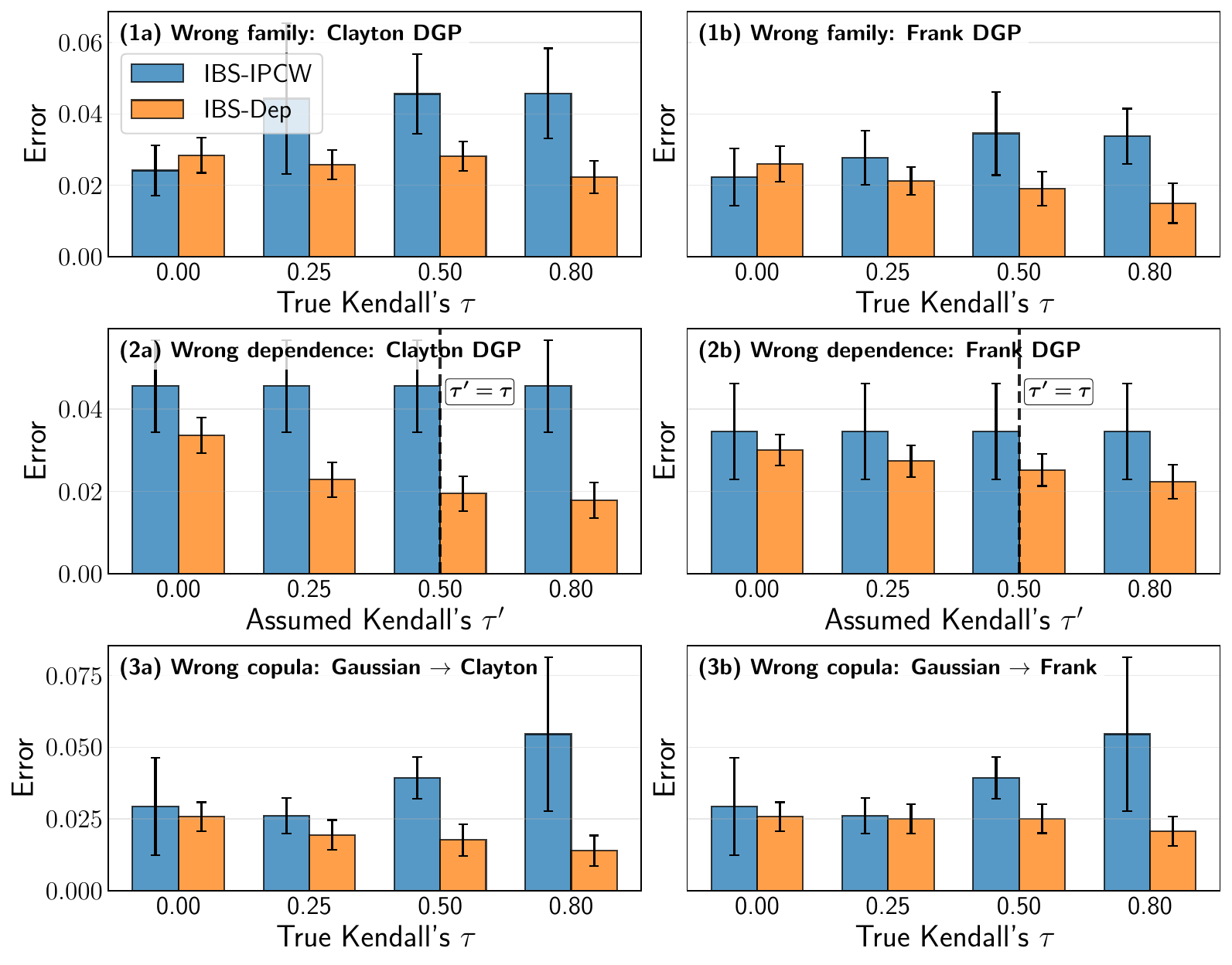}
\caption{Mean ($\pm$ SD) absolute estimation error of IBS-IPCW and IBS-Dep under copula misspecification, averaged over 10 experiments. The rows show (respectively) wrong family, wrong dependence strength, and wrong copula experiments. In the first two rows, columns separate Clayton and Frank data-generating copulas; in the final row, the Gaussian data-generating copula is evaluated using Clayton (resp., Frank) copulas. Censoring is held fixed at 50\%.}
\label{fig:synthetic_misspecification}
\end{figure}

\textbf{Misspecification of \(C_\theta\).}
Figure~\ref{fig:synthetic_misspecification} evaluates IBS-Dep when the assumed copula differs from the data-generating model. We consider three settings: (1) \emph{Wrong family}, where Clayton data are evaluated with Frank, or vice versa, at the same Kendall's $\tau$; (2) \emph{Wrong dependence}, where the family is correct but data generated with $\tau=0.5$ are evaluated using varying $\tau'$; and (3) \emph{Wrong copula}, where non-Archimedean Gaussian-copula data are evaluated using Clayton or Frank at the same Kendall's $\tau$. Across settings, IBS-Dep remains less sensitive to dependent censoring than IBS-IPCW, even with an imperfect assumed copula. However, the smallest error does not necessarily occur at the true $\tau$: IBS-Dep is slightly upward biased, and stronger assumed dependence can reduce this bias. We therefore interpret these results as a robustness analysis rather than a parameter-recovery diagnostic.

\begin{figure}
\centering
\includegraphics[width=1\linewidth]{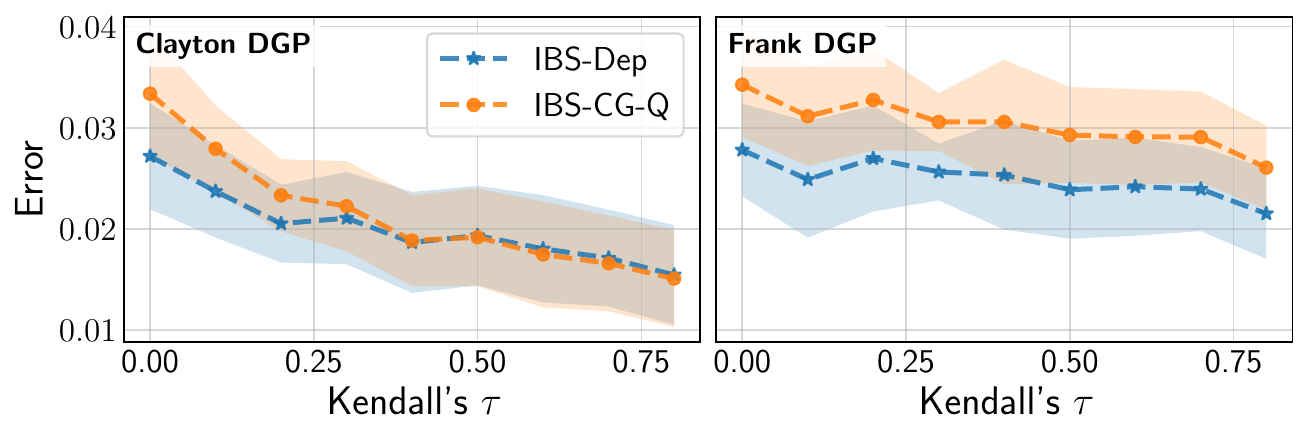}
\caption{Mean ($\pm$ SD) absolute estimation error of IBS-Dep and IBS-CG-Q under correctly specified copulas, shown separately for Clayton and Frank and averaged over 10 experiments. Censoring is held fixed at 50\%.}
\label{fig:synthetic_bg_vs_cgq}
\end{figure}

\textbf{Integrating over censored event-time uncertainty.}
IBS-Dep imputes each censored individual with a single CG-based margin time. To test whether this point estimate discards useful uncertainty, we also evaluate IBS-CG-Q, an ablation that integrates over the conditional event-time distribution after censoring. For evaluation time \(t\), define
\[
\widehat q_i^{\mathrm{CG}}(t)
=
\begin{cases}
\mathbbm{1}[t_i > t], & \delta_i = 1, \\
\Pr_{\mathrm{CG}}(E > t \mid E > t_i), & \delta_i = 0.
\end{cases}
\]
where \(\Pr_{\mathrm{CG}}\) denotes the conditional survival probability estimated by the CG estimator. The CG-Q Brier score replaces the binary survival status by \(\widehat q_i^{\mathrm{CG}}(t)\):
\begin{equation}
\label{eq:bs_cgq}
\begin{aligned}
\mathrm{BS}_{\mathrm{CG\text{-}Q}}(\mathcal{D}; \widehat{S}_E, t)
&= \frac{1}{N} \sum_{i=1}^N
\Bigl[
\widehat q_i^{\mathrm{CG}}(t)
\left(1-\SurvP{t}{\bm{x}_i}\right)^2 \\
&\quad+
\left(1-\widehat q_i^{\mathrm{CG}}(t)\right)
\SurvP{t}{\bm{x}_i}^2
\Bigr].
\end{aligned}
\end{equation}
The integrated version, IBS-CG-Q, is obtained by averaging Equation~\ref{eq:bs_cgq} over a prespecified interval $[t_1,t_2]$, analogously to IBS-Dep. Figure~\ref{fig:synthetic_bg_vs_cgq} compares IBS-Dep and IBS-CG-Q under correctly specified copulas. Overall, IBS-Dep gives slightly lower error than its conditional counterpart, although the two become nearly indistinguishable for stronger Clayton dependence. Both variants are upward biased across all dependence levels, with the largest bias at weak dependence and particularly under the Frank copula for IBS-CG-Q. As Kendall's $\tau$ increases, the upward bias decreases, so the apparent improvement at stronger dependence reflects reduced bias rather than an intrinsically easier censoring problem.


\begin{figure*}[!b]
\centering
\includegraphics[width=1\textwidth]{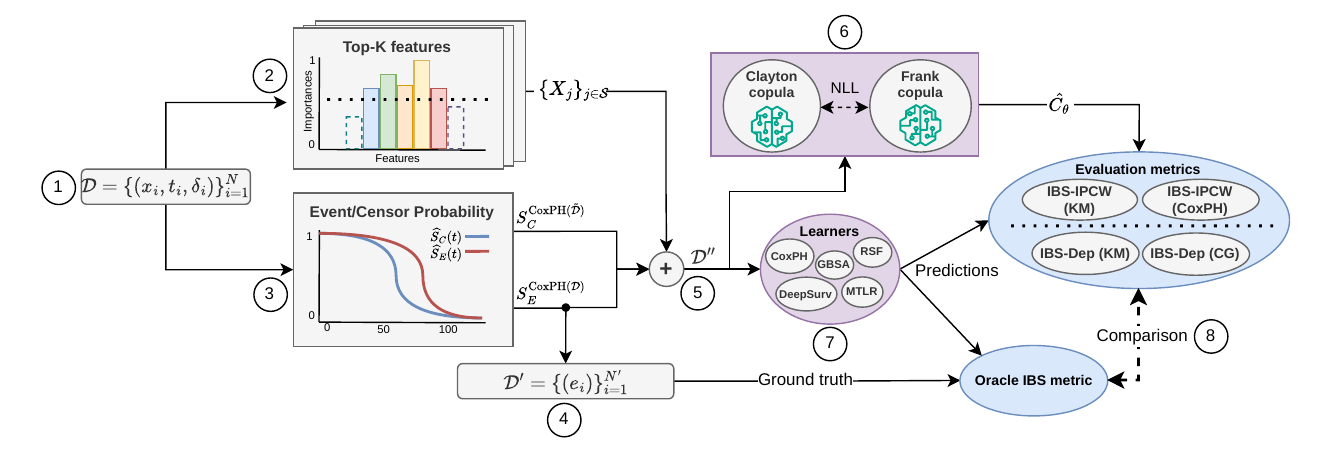}
\caption{Evaluation pipeline. (1) We begin with a real-world dataset $\mathcal{D}$. (2) A feature selection procedure (\eg top-$k$ selection) yields a reduced feature set. In parallel, (3) two CoxPH models are fitted to estimate the event- and censoring-time distributions, and (4) uncensored event times are sampled from the event distribution and stored as $\mathcal{D}'$ for oracle evaluation. (5) We construct a semi-synthetic dataset $\mathcal{D}''$ by combining the selected features with event and censoring times sampled from the fitted distributions. Next, (6) two Archimedean copula models are trained to estimate $\widehat{C}_\theta$, and (7) five survival learners are trained on $\mathcal{D}''$. Finally, (8) test-set predictions are evaluated using IBS-IPCW with KM or CoxPH censoring estimates and IBS-Dep with KM- or CG-based dependence estimates using $\widehat{C}_\theta$, and compared with the oracle IBS computed from $\mathcal{D}'$.}
\label{fig:evaluation_pipeline}
\end{figure*}

\section{Semi-synthetic experiments}
\label{sec:semisynthetic_experiments}

\subsection{Experimental setup}
\label{sec:semisynthetic_setup}

To evaluate IBS-Dep on real data, we require ground-truth event times, which are unavailable due to censoring. Following \citet{qi_effective_2023}, we construct semi-synthetic datasets (see Figure \ref{fig:evaluation_pipeline}) in which event and censoring mechanisms are learned from real data and used to resample outcomes while preserving the covariate structure. Dependent censoring is induced via targeted feature selection, creating settings where censoring depends on omitted information. The resulting dataset $\mathcal{D}''$ provides known event times for oracle evaluation while keeping realistic marginal behavior.

\textbf{Datasets} (Step 1 in Figure \ref{fig:evaluation_pipeline}): We apply the pipeline to 12 datasets: WHAS, METABRIC, Churn, GBSG, NACD, FLCHAIN, SUPPORT, Employee, MIMIC-IV, and 3 from SEER (death from brain, liver and stomach cancer). Table~\ref{tab:raw_datasets} summarizes the datasets; details are in Appendix~\ref{app:data_generation_and_processing}.

\begin{table}[!ht]
\caption{Overview of the raw datasets. $^\dagger$Number of features before (after) one-hot encoding.}
\label{tab:raw_datasets}
\centering
\resizebox{\columnwidth}{!}{%
\begin{tabular}{lrrr}
\toprule
Dataset & \#Instances & \#Features$^\dagger$ & \%Censored \\
\midrule
WHAS & 500 & 14 (14) & 57.0\% \\
GBSG & 686 & 8 (8) & 56.4\% \\
METABRIC & 1,902 & 9 (9) & 42.1\% \\
Churn & 1,958 & 12 (25) & 52.4\% \\
NACD & 2,396 & 48 (48) & 36.4\% \\
FLCHAIN & 7,871 & 9 (31) & 72.5\% \\
SUPPORT & 8,873 & 14 (27) & 32.0\% \\
Employee & 14,999 & 8 (17) & 76.2\% \\
MIMIC-IV & 38,520 & 91 (103) & 66.7\% \\
SEER-brain & 73,703 & 10 (10) & 40.1\% \\
SEER-liver & 82,841 & 14 (14) & 37.6\% \\
SEER-stomach & 100,360 & 14 (14) & 43.4\% \\
\bottomrule
\end{tabular}
}
\end{table}

\textbf{Feature selection} (Step 2 in Figure \ref{fig:evaluation_pipeline}): To induce dependent censoring, we use three strategies: (1) \emph{Original}, keeping all features; (2) \emph{Top-$k$}, selecting the most predictive features based on permutation importance~\citep{Breiman2001} using Harrell's C-index~\citep{harrell1996multivariable}; and (3) \emph{Random-\%}, selecting a random subset of the original features.

\textbf{Copula model} (Step 6 in Figure~\ref{fig:evaluation_pipeline}): To estimate dependence under censoring, we train Clayton and Frank copulas separately with linear Weibull proportional hazards models for the event-time and censoring-time marginals. In this semi-synthetic setting, where the true copula parameter is unknown, the dependence parameter $\theta$ is thus estimated from the data. Model parameters are learned by minimizing the penalized negative log-likelihood (NLL) under dependent censoring on the training set:
\begin{equation}
\label{eq:likelihood_dependent_main}
\begin{aligned}
&\ell_{\mathrm{pen}}(\theta;\mathcal{D})\\
&=
-\frac{1}{N}\sum_{i=1}^N
\Bigg\{
\delta_i
\Big[
\log f_E(t_i\mid\mathbf{x}_i)
+
\log \partial_{u_1}C_\theta(u_1,u_2)
\Big]
\\
&\qquad
+
(1-\delta_i)
\Big[
\log f_C(t_i\mid\mathbf{x}_i)
+
\log \partial_{u_2}C_\theta(u_1,u_2)
\Big]
\Bigg\}
\\
& \qquad
+
\lambda\theta^2 .
\end{aligned}
\end{equation}
Here, $u_1 = S_E(t_i \mid \mathbf{x}_i)$ and $u_2 = S_C(t_i \mid \mathbf{x}_i)$, where $C_\theta$ is an Archimedean copula and $\lambda=0.01$ is the regularization coefficient. The full derivation of the unpenalized likelihood is in Appendix~\ref{app:survival_likelihood}. We select the candidate model $\widehat{C}_\theta$ with the lowest validation penalized NLL. We optimize Equation~\ref{eq:likelihood_dependent_main} using Adam~\citep{KingBa15}, with separate learning rates for the Weibull marginal parameters and the copula model. The validation penalized NLL is also used for early stopping. Algorithm~\ref{alg:copula_estimation} summarizes the procedure. Only the selected copula family and parameter are used when computing IBS-Dep. See Appendix~\ref{app:implementation_details} for further implementation details.

\begin{algorithm}[!ht]
\caption{Copula estimation: PyTorch-like Pseudocode}
\label{alg:copula_estimation}
\begin{lstlisting}[style=pseudopy]
# model_e, model_c: Weibull models for E/C
# copula: Copula model with parameter theta
def find_copula(model_e, model_c, copula, d_train, d_val):

    opt = Adam(params=[model_e, model_c, copula])

    for epoch in range(n_epochs):
        for batch in DataLoader(d_train):
            train_nll = loss(model_e, model_c, batch, copula)
            opt.zero_grad()
            train_nll.backward()
            opt.step()

        # validate and track best loss
        val_losses = []
        for batch in DataLoader(d_val):
            val_losses.append(loss(model_e, model_c, batch, copula))
        val_nll = mean(val_losses)
        
        # check early stopping
        if theta_stabilized(copula) and no_val_improve(val_nll): break

    return copula

# Computes dependent NLL given a copula
def loss(model_e, model_c, batch, copula):
    T, E, X = batch.T, batch.E, batch.X
    s1 = model_e.survival(T, X) # S_E(T|X)
    s2 = model_c.survival(T, X) # S_C(T|X)
    f1 = model_e.pdf(T, X) # f_E(T|X)
    f2 = model_c.pdf(T, X) # f_C(T|X)

    # dependent likelihood via copula terms
    S = cat([s1, s2])
    p1 = log(f1) + log(copula.ccdf("u", S))
    p2 = log(f2) + log(copula.ccdf("v", S))

    reg = 0.01 * sum(copula.theta ** 2)
    ll = p1 * E + (1 - E) * p2
    return -mean(ll) + reg
\end{lstlisting}
\footnotesize \textbf{Notes:} 
\texttt{cat} concatenates matrices.
\texttt{log} calculates the natural logarithm.
\texttt{ccdf} computes a conditional CDF via differentiation.
\end{algorithm}

\textbf{Survival learners} (Step 7 in Figure \ref{fig:evaluation_pipeline}): We compare predictions from five survival learners: CoxPH~\citep{cox_regression_1972}, Gradient Boosting Survival Analysis (GBSA)~\citep{Ridgeway1999}, Random Survival Forests (RSF)~\cite{ishwaran_random_2008}, DeepSurv~\cite{katzman_deepsurv_2018} and Multi-Task Logistic Regression (MTLR)~\citep{NIPS2011_1019c809}. These learners are trained to predict survival curves $\widehat S_E(t \mid \bm{x}_i)$ for each instance $i$ in the test set. See Appendix~\ref{app:survival_learners} for implementation details and chosen hyperparameters.

We repeat the evaluation pipeline in Figure~\ref{fig:evaluation_pipeline} over 10 independent experiments. In each experiment, the semi-synthetic dataset $\mathcal{D}''$ is split into stratified training (70\%), validation (10\%), and test (20\%) sets (seeds 0–9). Missing values are imputed using training-set statistics (mean for continuous, mode for categorical variables). Numerical features are $z$-score normalized and categorical features one-hot encoded. To limit computational cost, large datasets ($N \geq 10{,}000$) are subsampled after preprocessing (Appendix~\ref{app:implementation_details}). As a sanity check, we plot empirical KM curves and event/censoring time distributions (see Figure~\ref{fig:semi_synth_distribution}).

\begin{table*}[!b]
\centering
\caption{Mean absolute estimation error of independent and proposed dependent IBS-based metrics across 12 datasets and 4 feature selection strategies. Results are averaged over 5 learners and 10 experiments. Relative (\%) change is shown against the KM-based IBS-IPCW baseline. \emph{IBS-IPCW (CoxPH)} denotes IPCW with a CoxPH censoring model, corresponding to conditional independent censoring. \emph{Dep (KM)} denotes the proposed metric using KM instead of CG as an ablation, and \emph{UW} denotes uncertainty weighting (see Section \ref{sec:definition}). Full results, including variances, are provided in Appendix~\ref{app:estimation_error}.}
\label{tab:estimation_error}
\resizebox{1\textwidth}{!}{
\begin{tabular}{llcccccccccccc|c}
& \multicolumn{1}{l}{Metric} & 
\rotatebox{90}{WHAS} &
\rotatebox{90}{METABRIC} &
\rotatebox{90}{Churn} &
\rotatebox{90}{GBSG} &
\rotatebox{90}{NACD} &
\rotatebox{90}{FLCHAIN} &
\rotatebox{90}{SUPPORT} &
\rotatebox{90}{Employee} &
\rotatebox{90}{MIMIC (IV)} &
\rotatebox{90}{SEER (brain)} &
\rotatebox{90}{SEER (liver)} &
\rotatebox{90}{SEER (stom.)} &
\rotatebox{90}{Average} \\
\midrule
\multirow{5}{*}{\rotatebox{90}{Original}}
& IPCW (KM) & 0.073 & 0.061 & 0.117 & \textbf{0.125} & 0.092 & 0.156 & 0.108 & 0.152 & 0.026 & 0.077 & 0.104 & 0.119 & 0.101 \\
& IPCW (CoxPH) & 0.106 & 0.051 & \textbf{0.115} & 0.129 & 0.148 & 0.162 & 0.109 & \textbf{0.100} & 0.039 & \textbf{0.055} & 0.097 & \textbf{0.095} & 0.100 \textcolor{improvGreen}{(-0.4\%)} \\
& Dep (KM) & \textbf{0.055} & 0.047 & 0.120 & 0.130 & 0.091 & 0.135 & 0.111 & 0.148 & \textbf{0.023} & 0.089 & 0.118 & 0.125 & 0.099 \textcolor{improvGreen}{(-1.4\%)} \\
& Dep (CG) & \textbf{0.055} & \textbf{0.047} & 0.120 & 0.126 & 0.091 & 0.135 & 0.109 & 0.154 & 0.024 & 0.084 & 0.106 & 0.112 & 0.097 \textcolor{improvGreen}{(-4.0\%)} \\
& Dep (CG)$^{\text{UW}}$ & 0.063 & 0.049 & 0.124 & 0.134 & \textbf{0.082} & \textbf{0.090} & \textbf{0.093} & 0.152 & 0.024 & 0.067 & \textbf{0.092} & 0.097 & \textbf{0.089} \textcolor{improvGreen}{(-11.8\%)} \\
\midrule
\multirow{5}{*}{\rotatebox{90}{Top-5}}
& IPCW (KM) & 0.071 & 0.062 & 0.121 & \textbf{0.124} & 0.103 & 0.133 & 0.109 & 0.155 & 0.027 & 0.076 & 0.109 & 0.124 & 0.101 \\
& IPCW (CoxPH) & 0.065 & 0.061 & \textbf{0.114} & 0.126 & 0.098 & 0.138 & 0.109 & \textbf{0.103} & 0.027 & 0.083 & 0.109 & 0.128 & 0.097 \textcolor{improvGreen}{(-4.4\%)} \\
& Dep (KM) & \textbf{0.055} & 0.048 & 0.120 & 0.130 & 0.098 & 0.108 & 0.112 & 0.149 & 0.026 & 0.090 & 0.120 & 0.126 & 0.098 \textcolor{improvGreen}{(-2.8\%)} \\
& Dep (CG) & \textbf{0.055} & \textbf{0.048} & 0.119 & 0.126 & 0.097 & 0.108 & 0.109 & 0.156 & 0.027 & 0.085 & 0.111 & 0.116 & 0.096 \textcolor{improvGreen}{(-4.9\%)} \\
& Dep (CG)$^{\text{UW}}$ & 0.065 & 0.049 & 0.125 & 0.134 & \textbf{0.089} & \textbf{0.066} & \textbf{0.094} & 0.154 & \textbf{0.025} & \textbf{0.068} & \textbf{0.096} & \textbf{0.101} & \textbf{0.089} \textcolor{improvGreen}{(-12.2\%)} \\
\midrule
\multirow{5}{*}{\rotatebox{90}{Top-10}}
& IPCW (KM) & 0.071 & 0.061 & 0.119 & \textbf{0.125} & 0.098 & 0.143 & 0.109 & 0.152 & 0.026 & 0.077 & 0.103 & 0.119 & 0.100 \\
& IPCW (CoxPH) & 0.082 & 0.051 & \textbf{0.111} & 0.130 & 0.093 & 0.150 & 0.109 & \textbf{0.101} & 0.031 & \textbf{0.055} & 0.097 & 0.099 & 0.092 \textcolor{improvGreen}{(-7.9\%)} \\
& Dep (KM) & \textbf{0.054} & 0.047 & 0.121 & 0.131 & 0.095 & 0.123 & 0.111 & 0.147 & \textbf{0.023} & 0.090 & 0.118 & 0.124 & 0.099 \textcolor{improvGreen}{(-1.4\%)} \\
& Dep (CG) & \textbf{0.054} & \textbf{0.047} & 0.121 & 0.126 & 0.095 & 0.123 & 0.109 & 0.153 & 0.024 & 0.085 & 0.107 & 0.111 & 0.096 \textcolor{improvGreen}{(-4.0\%)} \\
& Dep (CG)$^{\text{UW}}$ & 0.063 & 0.049 & 0.125 & 0.134 & \textbf{0.086} & \textbf{0.078} & \textbf{0.093} & 0.152 & 0.025 & 0.068 & \textbf{0.093} & \textbf{0.097} & \textbf{0.088} \textcolor{improvGreen}{(-11.7\%)} \\
\midrule
\multirow{5}{*}{\rotatebox{90}{Rand. 25\%}}
& IPCW (KM) & 0.090 & 0.061 & 0.129 & 0.153 & 0.109 & 0.140 & 0.142 & 0.161 & \textbf{0.016} & 0.111 & 0.128 & 0.132 & 0.114 \\
& IPCW (CoxPH) & 0.091 & 0.056 & 0.122 & 0.157 & 0.111 & 0.147 & 0.142 & \textbf{0.110} & 0.017 & 0.111 & 0.197 & 0.133 & 0.116 \textcolor{improvRed}{(+1.6\%)} \\
& Dep (KM) & \textbf{0.081} & 0.049 & 0.126 & 0.153 & 0.101 & 0.140 & 0.145 & 0.137 & 0.031 & 0.113 & 0.127 & 0.135 & 0.111 \textcolor{improvGreen}{(-2.6\%)} \\
& Dep (CG) & \textbf{0.081} & \textbf{0.049} & \textbf{0.117} & \textbf{0.147} & 0.099 & 0.140 & 0.141 & 0.120 & 0.032 & 0.102 & 0.109 & 0.120 & 0.105 \textcolor{improvGreen}{(-8.4\%)} \\
& Dep (CG)$^{\text{UW}}$ & 0.095 & 0.050 & 0.131 & 0.161 & \textbf{0.091} & \textbf{0.032} & \textbf{0.125} & 0.147 & 0.021 & \textbf{0.089} & \textbf{0.098} & \textbf{0.106} & \textbf{0.096} \textcolor{improvGreen}{(-16.4\%)} \\
\bottomrule
\end{tabular}
}
\end{table*}

\begin{table}[!ht]
\centering
\caption{Ranking performance of the independent and proposed dependent metrics across 12 datasets and 4 feature selection strategies. The results show how often each metric recovered at least two of the top three survival learners according to the oracle IBS, across 10 experiments per feature selection strategy. Each entry is therefore reported out of 40. Higher is better.}
\label{tab:ranking_results}
\resizebox{1\columnwidth}{!}{
\begin{tabular}{l|ccccc}
\toprule
Dataset
& \shortstack{IPCW\\(KM)}
& \shortstack{IPCW\\(CoxPH)}
& \shortstack{Dep\\(KM)}
& \shortstack{Dep\\(CG)}
& \shortstack{Dep\\(CG)$^{\text{UW}}$} \\
\midrule
WHAS
& 38/40
& 39/40
& \textbf{40/40}
& \textbf{40/40}
& 39/40 \\

METABRIC
& \textbf{40/40}
& \textbf{40/40}
& 39/40
& 39/40
& \textbf{40/40} \\

Churn
& \textbf{37/40}
& \textbf{37/40}
& 34/40
& 29/40
& 32/40 \\

GBSG
& \textbf{40/40}
& \textbf{40/40}
& 38/40
& 37/40
& 35/40 \\

NACD
& 39/40
& \textbf{40/40}
& 38/40
& 36/40
& 38/40 \\

FLCHAIN
& \textbf{39/40}
& \textbf{39/40}
& 18/40
& 18/40
& 35/40 \\

SUPPORT
& \textbf{37/40}
& 36/40
& 35/40
& 34/40
& 31/40 \\

Employee
& 33/40
& \textbf{35/40}
& 12/40
& 11/40
& 13/40 \\

MIMIC (IV)
& \textbf{33/40}
& 29/40
& 1/40
& 1/40
& 13/40 \\

SEER (brain)
& 37/40
& \textbf{38/40}
& \textbf{38/40}
& 36/40
& 36/40 \\

SEER (liver)
& \textbf{39/40}
& \textbf{39/40}
& 38/40
& 30/40
& 35/40 \\

SEER (stom.)
& 36/40
& 38/40
& \textbf{40/40}
& 35/40
& 35/40 \\
\bottomrule
\end{tabular}%
}
\end{table}

\subsection{Results and takeaways}

\textbf{Estimation error.} Table~\ref{tab:estimation_error} reports the mean estimation error on the semi-synthetic datasets, measured as the absolute deviation from the oracle IBS computed using the true event times. On average, the CG-based estimator reduces error by 4-8\%, while the uncertainty-weighted variant achieves the largest reductions, at 12-16\%. The relative improvement becomes more pronounced when important features are omitted (\eg Random 25\%). This does not imply that the KM estimate of the censoring distribution becomes more biased; however, the resulting Brier-score bias can still depend on the learner's prediction errors. When informative features are removed, prediction errors increase; under dependent censoring, IPCW applies misspecified inverse weights to these larger squared errors, amplifying the resulting bias. In addition, with weaker (random) covariates, the marginal model explains less variation in $E$, leaving more unexplained (residual) dependence between $E$ and $C$. The CoxPH-based IPCW estimator also depends strongly on the selected feature set: it \emph{reduces} error by up to 8\% for the Original, Top-5, and Top-10 strategies, but \emph{increases} error by 2\% under the Random 25\% strategy. This highlights a key limitation of conditional IPCW, since an uninformative feature set can yield a misspecified censoring model and invalidate conditional independent censoring. The ablation study highlights the importance of the CG estimator. Replacing censored event times with margin times derived from KM yields only modest gains of 1-3\% over IBS-IPCW. In contrast, the CG estimator estimates the marginal event-time distribution while accounting for dependence between event and censoring times through the specified copula model, and uses this distribution to compute the conditional margin time for censored instances. This increases the reduction to 4-8\% across feature selection strategies, indicating that accounting for dependent censoring when estimating the event-time margin provides substantial improvements over KM-based margin imputation.

\textbf{Ranking.} Table~\ref{tab:ranking_results} reports the ranking experiment, which evaluates whether a censored metric identifies the same top-performing models as the oracle IBS. We first compute the oracle IBS for each model (\ie without censoring) and identify the top-3 models, then assess whether each censored metric recovers at least two of them. IBS-IPCW preserves the oracle top-3 most consistently overall, while the dependent metrics perform comparably on several datasets but worse on FLCHAIN, Employee, and MIMIC-IV. Thus, although the copula-based IBS substantially reduces absolute estimation error under dependent censoring, this does not necessarily translate into better separation between closely performing models. IBS-IPCW may therefore remain reasonable for within-dataset ranking when its bias affects competing models similarly. However, it does not estimate the true IBS under dependent censoring and may give unfair rankings when models differ in how they handle censoring.

\textbf{Computational analysis.} The primary computational cost lies in copula fitting, which jointly optimizes the marginal survival models and copula parameter with model selection over candidate families (Clayton and Frank). Empirically, this required 5-75 minutes, depending on dataset size (Appendix~\ref{app:computational_analysis}). Once estimated, IBS-Dep adds negligible cost relative to IPCW. The incremental peak PyTorch GPU memory allocation is modest ($\leq 11$ MiB), indicating that the method requires little additional GPU memory.

\section{Discussion}

\subsection{Main findings and implications}

Our results show that accounting for marginal dependent censoring at evaluation time can meaningfully change how we judge model performance. Across synthetic and semi-synthetic experiments, IBS-Dep consistently reduced bias relative to IBS-IPCW when event and censoring times were dependent. The copula-based estimator achieved average reductions of 4-8\%, while the uncertainty-weighted variant yielded improvements of 12-16\%. These effects were observed across 12 generated survival datasets with sample sizes from 500 to 10{,}000, feature dimensions from 14 to 91, and censoring rates between 32\% and 76\%. Overall, this suggests that dependent censoring is not a rare edge case, but a relevant trait of many real-world datasets. Although IBS-Dep reduced estimation error, it did not consistently improve the ranking of models, because bias reduction does not necessarily improve separation between models.

Most prior work concerning dependent censoring has concentrated on \emph{model estimation}. Copula-based modeling approaches~\citep{Foomani2023,zhang2024deep,liu2025hacsurv} explicitly model the joint distribution of event and censoring times to capture their dependence. However, these models are typically evaluated using metrics such as Harrell's C-index or Brier score with IPCW, both of which assume independent censoring. This creates a mismatch: models are designed for dependence, but evaluation still assumes independence. As a result, reported performance in dependent censoring settings may itself be biased. Our work shifts the focus to \emph{model evaluation}, introducing a flexible copula-based extension of the Brier score.

\subsection{Limitations and future work}

Our work also points to several directions for future research. More broadly, the framework introduced here opens the door to dependence-aware versions of other common evaluation metrics. Widely used measures -- such as Harrell's C-index, Uno's C-index~\citep{Uno2011}, and D-calibration~\citep{haider_effective_2020} -- rely on independence assumptions. Extending copula-based corrections to concordance and calibration measures is a natural next step toward having metrics that all support dependent censoring.

At the same time, several limitations remain. First, we assume a stable dependence structure; under temporal distribution shift, the copula parameter must be re-estimated, suggesting the need for adaptive or time-varying dependence models. Second, the true copula is not identifiable in real-world survival datasets because for each individual only one of the competing times is observed~\citep{Tsiatis1975}. Some misspecification is therefore unavoidable. Third, likelihood-based estimation can be unstable in small samples, particularly under dependent censoring, because weak identification of the copula parameter can make its estimate highly sensitive to sampling variation. These limitations motivate reporting sensitivity analyses across plausible copula families and dependence strengths. Agreement across specifications would strengthen confidence in the evaluation, whereas disagreement would reveal that model comparisons are sensitive to assumptions about the censoring mechanism.

\subsection{User guide}

We conclude with a brief guide on how to use IBS-Dep. The central choice is the dependence model used to approximate the relationship between event and censoring times. In standard right-censored data, we observe $t_i=\min(e_i,c_i)$ and $\delta_i=\mathbbm{1}(e_i\leq c_i)$, but only the smaller of $e_i$ and $c_i$. Thus, the joint distribution of the population-level variables $E$ and $C$, and hence their true dependence structure, is generally not identifiable without additional assumptions. We therefore propose not attempting to recover the true latent copula, but instead using a quantitative model comparison to determine whether an Archimedean copula fits the data better than independence.

In practice, one specifies candidate Archimedean copula families and parametric marginal models for $E$ and $C$, whose parameters are estimated by minimizing the dependent NLL in Equation~\ref{eq:likelihood_dependent_main}. As a baseline, one should fit the independence copula using the same marginal specification. The model with the best validation likelihood is selected; if none of the candidate dependent copulas improves upon the independence copula, one should default to independence. Algorithm~\ref{alg:copula_estimation} summarizes the copula estimation procedure. The selected copula is treated as a working model for the censoring mechanism, and the CG estimate of the event-time distribution is used to impute margin times for censored instances. The selected family is not claimed to be the true latent copula, but a likelihood-based approximation of the dependence structure between event and censoring times.

\begin{acknowledgements}
This research received support from the Natural Sciences and Engineering Research Council of Canada (NSERC), the Canadian Institute for Advanced Research (CIFAR), and the Alberta Machine Intelligence Institute (Amii).
\end{acknowledgements}

\bibliography{references}

\appendix
\onecolumn

\numberwithin{figure}{section}
\numberwithin{table}{section}
\numberwithin{algorithm}{section}

\title{Overcoming Dependent Censoring in the Evaluation of Survival Models\\(Supplementary Material)}
\maketitle

\section{Censoring assumptions}
\label{app:censoring_assumptions}

\textbf{Random censoring}. Instances censored at time \( t \) are representative of all instances still at risk at $t$, in terms of their survival experience~\citep[Ch. 1]{kleinbaum2012survival}. In a medical study, we would assume that the death rate of censored individuals is the same as that of uncensored individuals who remain in the risk set. Formally, $E \independent C$, where $E$ is the event time and $C$ is the censoring time, meaning $\Pr(E > t \mid C = t) = \Pr(E > t)$ for all $t$.

\textbf{Independent Censoring}. In the absence of competing risks (\ie when no alternative event precludes the main event), within any subgroup of interest (\eg a smoker and a non-smoker group), instances censored at time $t$ are representative of those who remain at risk at time $t$ with respect to their survival experience~\citep{kleinbaum2012survival}. This is also known as conditional independent censoring, since we assume that the observed covariates account for the relationship between event and censoring. Thus, once we condition on the covariates, the censoring time is assumed to be independent of the event time. Formally, $E \independent C \mid X$, where $X$ are the covariates, meaning $\Pr(E > t \mid C = t, X) = \Pr(E > t \mid X)$ for all $t$. Independent censoring is a less restrictive assumption than random censoring on the data-generating process.

\textbf{Dependent Censoring}. In the absence of competing risks, within any subgroup of interest, instances censored at time $t$ are \emph{not} representative of those who remain at risk at time $t$ with respect to their survival experience~\citep{Foomani2023,lillelund_stop_2026}. Equivalently, censoring is said to be dependent when, even after conditioning on the observed covariates, the censoring mechanism remains related to the event process. Thus, the observed covariates do not fully account for the relationship between $E$ and $C$, and the censoring time is not independent of the event time given the covariates.

\section{Copulas for survival analysis}\label{app:copulas_for_survival}

Bivariate survival analysis can be viewed as a \textit{competing risks} problem, where the occurrence of an event (\eg death) and censoring (\eg dropout) are mutually exclusive. \citet{Tsiatis1975} referred to this as the \textit{identifiability issue}; competing risks data do not allow one to observe two event times simultaneously (one event censors the other), and hence, the data may not identify the dependence structure between event times. \citet{Zheng1995} gave a partial solution to the identifiability problem of dependent censoring by an assumed \textit{copula} between two event times. A copula links two or more random variables by specifying their dependence structure~\citep{Emura2018}. A mathematician, Abe Sklar, first used the word "copula" in his study of probabilistic metric spaces~\citep{Sklar1959}. In his work, he gave a mathematical definition of copulas and established the most fundamental theorem about copulas, known as Sklar's theorem~\citep{Sklar1959}. It demonstrated that any joint cumulative distribution
function can be written in terms of a copula over the quantiles of its marginal cumulative distribution functions. Among the various types of copulas, the Archimedean copula is the most widely used, and enables the modeling of dependencies in arbitrarily high dimensions using a single parameter. This family of copulas is defined by a convex generator function, $\varphi_\theta: [0, 1] \to [0, \infty]$, which is continuous, strictly decreasing, and satisfies the boundary conditions $\varphi_\theta(0) = \infty$ and $\varphi_\theta(1) = 0$~\citep{Emura2018}. Given event time $E$ and censoring time $C$, their joint survival function under an Archimedean copula can be written as:
\begin{equation}
\Pr(E > t, C > c) = \varphi_\theta^{-1} \left[
\varphi_\theta\!\left(S_E(t)\right) +
\varphi_\theta\!\left(S_C(c)\right) \right],
\end{equation}
where $\varphi_\theta$ is the generator function, and
$S_E(t)=\Pr(E>t)$ and $S_C(c)=\Pr(C>c)$ are the marginal survival functions. In the bivariate case, $\varphi_\theta$ describes the dependence structure (\ie correlation) between two random variables, \eg $E$ and $C$, and the (rank) correlation can be described with Kendall's tau ($\tau$):
\begin{equation}
\begin{split}
\tau = &\quad \Pr{\ (t_2 - t_1)(c_2 - c_1) \geq 0 } - \Pr{\ (t_2 - t_1)(c_2 - c_1) < 0 },
\end{split}
\end{equation}
\noindent where ($t_1$, $c_1$) and ($t_2$, $c_2$) are sampled from a (survival) copula. The Clayton~\citep{clayton1978model} and Frank~\citep{frank1979simultaneous} copulas are two common Archimedean copulas in survival literature. Within these families, the copula $C_{\theta}$ is parameterized by $\theta$, interpreted as the degree of dependence between $E$ and $C$ in the bivariate case. When $\theta \approx 0$, it yields the independence copula for Clayton and Frank copulas; thus, a larger value of $\theta$ implies greater dependence between the marginal distributions. Formally, we write $C(u_1, \ldots, u_d) : [0, 1]^d \rightarrow [0, 1]$, which is a $d$-dimensional copula (\ie where $d$ is the number of variables) with $u_1, \ldots, u_d$ uniform marginal probabilities and the following properties~\citep{Nelsen2006}:

1. \textbf{Groundedness}: If there exists an \( i \in \{1, \dots, d\} \) such that \( u_i = 0 \), then  
\[
C(u_1, \dots, u_d) = 0.
\]
2. \textbf{Uniform Margins}: For all \( i \in \{1, \dots, d\} \), if \( \forall j \neq i \Rightarrow u_j = 1 \), then  
\[
C(u_1, \dots, u_d) = u_i.
\]
3. \textbf{$d$-Increasingness}: For all \( u = (u_1, \dots, u_d) \) and \( v = (v_1, \dots, v_d) \) where \( u_i < v_i \) for all \( i = 1, \dots, d \), the following holds:  
\[
\sum_{\ell \in \{0,1\}^d} (-1)^{\ell_1 + \dots + \ell_d} C\left( u_1^{\ell_1} v_1^{1-\ell_1}, \dots, u_d^{\ell_d} v_d^{1-\ell_d} \right) \geq 0.
\]
Let $E$ be the event distribution, $C$ the censoring distribution, $\bm{x}$ a vector of covariates, and $S_E(t\mid\bm{x}) = \Pr(E > t\mid\bm{x})$ and $S_C(c\mid\bm{x}) = \Pr(C > c\mid\bm{x})$ the marginal survival functions given $\bm{x}$. We can then define a bivariate (survival) copula $C_{\theta}$~\citep{Nelsen2006} that describes the degree of dependence between $E$ and $C$ as follows:
\begin{equation}
\label{eq:survival_copula}
\Pr(E > t, C > c \mid \bm{x}) =
C_{\theta}\!\left(
S_E(t \mid \bm{x}),
S_C(c \mid \bm{x})
\right).
\end{equation}
Kendall's tau ($\tau$) can be defined as:
\begin{equation}
\begin{split}
\tau = &\quad \Pr{\ (t_2 - t_1)(c_2 - c_1) \geq 0\mid \bm{x}\ } - \Pr{\ (t_2 - t_1)(c_2 - c_1) < 0 \mid \bm{x}\ },
\end{split}
\end{equation}
\noindent where ($t_1$, $c_1$) and ($t_2$, $c_2$) are sampled from the model~\eqref{eq:survival_copula}. Kendall's $\tau$ can be solely expressed as a function of the copula $C_\theta$ by~\citep{Emura2018}:
\begin{equation}
\tau_{\theta} = 4\int_0^1 \int_0^1 C_\theta(u, v) , C_\theta(du, dv),
\end{equation}
\noindent which implies that Kendall's $\tau$ does not depend on the marginal survival functions. An Archimedean copula is a type of copula that is defined using a generator function, which allows for a flexible way to model dependence structures between random variables, \eg $E$ and $C$. Such a copula is defined as~\citep{Emura2018}:
\begin{equation}
C_{\theta}(u,v) =
\varphi_\theta^{-1}\left(
\varphi_\theta(u)+\varphi_\theta(v)
\right).
\end{equation}
\noindent where the function $\varphi_\theta : [0, 1] \rightarrow [0, \infty]$ is called a generator of the copula, which is continuous and strictly decreasing from $\varphi_\theta(0) = \infty$ to $\varphi_\theta(1) = 0$. The following Archimedean copulas are considered in this work:

\begin{itemize}
\item \textbf{The independence copula:}
\begin{equation}
C(u, v) = uv.
\label{eq:indep-copula}
\end{equation}

\item \textbf{The Clayton copula~\citep{clayton1978model}:}
\begin{equation}
C_\theta(u, v) = \left( u^{-\theta} + v^{-\theta} - 1 \right)^{-1/\theta},
\quad \theta \geq 0.
\label{eq:clayton-copula}
\end{equation}

\item \textbf{The Frank copula~\citep{frank1979simultaneous}:}
\begin{equation}
C_\theta(u, v) = -\frac{1}{\theta} \log \left[ 1 +
\frac{(\exp(-\theta u) - 1)(\exp(-\theta v) - 1)}{e^{-\theta} - 1} \right],
\quad \theta \neq 0.
\label{eq:frank-copula}
\end{equation}
\end{itemize}

\section{Survival likelihood}
\label{app:survival_likelihood}

In the standard survival analysis framework, we define the survival time $E$ and censoring time $C$ such that:
\begin{align*}
E \ = \ t_i \ \text{and} \ C \ \geq \ t_i &\qquad \text{if} \ \delta_i \ = \ 1, \\
E \ > \ t_i \ \text{and} \ C \ = \ t_i &\qquad \text{if} \ \delta_i \ = \ 0.
\end{align*}
Combining these cases, the likelihood for the $i$-th observation is:
\begin{equation}
\label{eq:likelihood_full}
\mathcal{L}_i = \Pr\left(E = t_i, C \geq t_i \,\middle|\, x_i\right)^{\delta_i} \; \Pr\left(E \geq t_i, C = t_i \,\middle|\,  x_i\right)^{1 - \delta_i}
\end{equation}
Under the assumption of independent censoring~\citep{Emura2018}, we have:
\begin{align}\label{eq:likelihood_independent}
\mathcal{L}_i &= \left[ \Pr\left(E = t_i \,\middle|\,  x_i\right) \, \Pr\left(C \geq t_i \,\middle|\,  x_i\right) \right]^{\delta_i} \left[ \Pr\left(E \geq t_i \,\middle|\,  x_i\right) \, \Pr\left(C = t_i \,\middle|\,  x_i\right) \right]^{1 - \delta_i} \notag \\
&= \left[ f_E\left(t_i \,\middle|\,  x_i\right) \, S_C\left(t_i \,\middle|\,  x_i\right) \right]^{\delta_i} \left[ S_E\left(t_i \,\middle|\,  x_i\right) \, f_C\left(t_i \,\middle|\,  x_i\right) \right]^{1 - \delta_i} \notag \\
&= \left[ f_E\left(t_i \,\middle|\,  x_i\right)^{\delta_i} S_E\left(t_i \,\middle|\,  x_i\right)^{1 - \delta_i} \right] \left[ f_C\left(t_i \,\middle|\,  x_i\right)^{1 - \delta_i} S_C\left(t_i \,\middle|\,  x_i\right)^{\delta_i} \right]
\end{align}
where $S_E(t \mid x_i) = \Pr(E \geq t \mid x_i)$, $f_E(t \mid x_i) = -dS_E(t \mid x_i)/dt$, $S_C(t \mid x_i) = \Pr(C \geq t \mid x_i)$, and $f_C(t \mid x_i) = -dS_C(t \mid x_i)/dt$.  
Under the non-informative censoring assumption, the term $f_C(t_i \mid x_i)^{1 - \delta_i} S_C(t_i \mid x_i)^{\delta_i}$ is unrelated to the likelihood for the event times and can simply be ignored~\citep{Emura2018}. Therefore, the likelihood function can be written as:
\begin{equation}\label{eq:likelihood_noninformative}
\mathcal{L} = \prod_{i=1}^{N} f_E(t_i \mid \bm{x}_i)^{\delta_i} S_E(t_i \mid \bm{x}_i)^{1 - \delta_i}, 
\end{equation}
where $f_E(t_i \mid x_i)$ is the density function of the event time and $S_E(t_i \mid x_i)$ is the survival function of the event time. Note how the censoring terms $f_C(t_i)$ and $S_C(t_i)$ are absent in \eqref{eq:likelihood_noninformative}. If the event is observed, the likelihood contribution is the probability density of the event occurring at that time. If the observation is censored, the likelihood contribution is the probability of surviving beyond that time. The overall likelihood is the product of these contributions across all instances.

However, if censoring is dependent, we can no longer rely on this clean decomposition of the likelihood as in Equation \ref{eq:likelihood_independent}. Instead, given some Archimedean copula (\eg Clayton or Frank), we can rewrite Equation \ref{eq:likelihood_independent} as the log-likelihood under dependent censoring~\citep{Foomani2023}:
\begin{align}
\label{eq:likelihood_dependent}
\ell(\theta; \mathcal{D}) = \sum_{i=1}^N \delta_{i} \; \log \left[f_{E}(t_i|\bm{x}_{i})\right] \; 
+ \; \delta_{i} \; \text{log} \left[\frac{\partial}{\partial u_1}C(u_1, u_2)\bigg\rvert^{u_1 = S_{E}(t_i \rvert \bm{x}_{i})}_{u_2 = S_{C}(t_i \rvert \bm{x}_{i})}\right] \notag \\
+ \; (1 - \delta_{i}) \; \text{log} \left[f_{C}(t_i|\bm{x}_{i})\right] + (1 - \delta_{i}) \; \log \left[\frac{\partial}{\partial u_2}C(u_1, u_2)\bigg\rvert^{u_1 = S_{E}(t_i \rvert \bm{x}_{i})}_{u_2 = S_{C}(t_i \rvert \bm{x}_{i})}\right]
\end{align}
In this expression, the first term represents the log-likelihood of observing the event at time $t_i$. The second term captures the conditional probability of the censoring time occurring after the event time, given that the event occurred at $t_i$. By symmetry, the third and fourth terms mirror these quantities but with respect to the censoring time. Although Equation \ref{eq:likelihood_dependent} may appear complex, the partial derivatives of any Archimedean copula have closed-form expressions, which allows us to maximize the log-likelihood using gradient descent.

\section{Theoretical properties and derivations}
\label{app:theoretical_properties}

This appendix develops the theoretical foundations of the proposed CG margin-time estimator under dependent censoring. We first derive the KM and CG estimators to establish the survival estimation framework under independent and dependent censoring, respectively, which forms the basis for the subsequent margin-time analysis. We then recall the asymptotic target of the CG survival estimator under possible copula misspecification and introduce the margin-time functional and its plug-in CG estimator.

\subsection{The Kaplan-Meier estimator}
\label{app:km_estimator}

The following derivation follows the presentation of \citet{Emura2018}, generalized to accommodate tied event times.

Recall that $(t_i,\delta_i)$, $i=1,\ldots,N$, denotes survival data
without covariates, where $t_i=\min(e_i,c_i)$ and
$\delta_i=\mathbbm{1}[e_i\leq c_i]$. Let
$u_1<\cdots<u_J$ denote the distinct observed event times. For each
$u_j$, define
\[
d_j
=
\sum_{i=1}^{N}
\mathbbm{1}[t_i=u_j,\delta_i=1]
\]
as the number of events at $u_j$, and
\[
n_j
=
\sum_{i=1}^{N}
\mathbbm{1}[t_i\geq u_j]
\]
as the number of instances at risk immediately before $u_j$.

We represent $S_E(t)$ as a decreasing step function whose jumps occur at the observed event times. Its value at time $t$ can therefore be written as
\begin{align}
\label{eq:km1}
S_E(t)
&=
\prod_{u_j\leq t}
\frac{S_E(u_j)}{S_E(u_j^-)}
\\
&=
\prod_{u_j\leq t}
\left(
1-
\frac{\Pr(E=u_j)}
{\Pr(E\geq u_j)}
\right),
\end{align}
where $S_E(u_j^-)$ denotes the survival probability immediately before the jump at $u_j$.

Suppose that $E$ and $C$ are independent. Then,
\begin{align}
\label{eq:k2}
S_E(t)
&=
\prod_{u_j\leq t}
\left(
1-
\frac{\Pr(E=u_j,C\geq u_j)}
{\Pr(E\geq u_j,C\geq u_j)}
\right)
\\
&=
\prod_{u_j\leq t}
\left(
1-
\frac{
\Pr\!\left(\min(E,C)=u_j,E\leq C\right)
}{
\Pr\!\left(\min(E,C)\geq u_j\right)
}
\right).
\end{align}

Replacing each probability ratio by its empirical estimate gives
\begin{align}
\label{eq:km3}
\widehat{S}_E(t)
&=
\prod_{u_j\leq t}
\left(
1-
\frac{
\sum_{i=1}^{N}
\mathbbm{1}[t_i=u_j,\delta_i=1]
}{
\sum_{i=1}^{N}
\mathbbm{1}[t_i\geq u_j]
}
\right)
\\
&=
\prod_{u_j\leq t}
\left(
1-\frac{d_j}{n_j}
\right),
\qquad
0\leq t\leq t_{\max}.
\end{align}
Thus, all events tied at $u_j$ are incorporated through the common
factor $1-d_j/n_j$. The empty product is defined as one, and the
estimator is undefined for $t>t_{\max}$.

\subsection{The Copula-Graphic estimator}
\label{app:cg_estimator}

The following derivation is based on the Copula-Graphic estimator introduced by \citet{Zheng1995} and its explicit Archimedean representation derived by \citet{Rivest2001}, generalized here to accommodate tied event times.

Consider the Archimedean copula model
\begin{equation}
\label{eq:cg1}
\Pr(E>t,C>c)
=
\varphi_\theta^{-1}
\left[
\varphi_\theta\!\left(S_E(t)\right)
+
\varphi_\theta\!\left(S_C(c)\right)
\right],
\end{equation}
where $\varphi_\theta:[0,1]\rightarrow[0,\infty]$ is a continuous,
strictly decreasing generator satisfying
$\varphi_\theta(0)=\infty$ and $\varphi_\theta(1)=0$~\citep{Emura2018}.
The marginal survival functions are
$S_E(t)=\Pr(E>t)$ and $S_C(t)=\Pr(C>t)$.

Let $u_1<\cdots<u_J$ denote the distinct observed event times, and let
$d_j$ and $n_j$ be defined as in Section~\ref{sec:background_and_related_work}.
We treat all events tied at $u_j$ as a single jump of the event survival
function. When events and censorings occur at the same observed time,
we use the event-before-censoring convention: the event jump is applied
before any censoring removals at that time.

Immediately before the event jump at $u_j$, the empirical joint
survival probability is
\[
\widehat{\Pr}(E\geq u_j,C\geq u_j)
=
\frac{n_j}{N}.
\]
Immediately after removing the $d_j$ events at $u_j$, but before
removing any censorings tied at that time, it is
\[
\frac{n_j-d_j}{N}.
\]

Because only the event survival function changes across this event
jump, the copula representation gives
\begin{equation}
\label{eq:cg2}
\varphi_\theta\!\left(\frac{n_j-d_j}{N}\right)
=
\varphi_\theta\!\left(\widehat{S}_E(u_j)\right)
+
\varphi_\theta\!\left(\widehat{S}_C(u_j^-)\right),
\end{equation}
whereas immediately before the jump,
\begin{equation}
\label{eq:cg3}
\varphi_\theta\!\left(\frac{n_j}{N}\right)
=
\varphi_\theta\!\left(\widehat{S}_E(u_j^-)\right)
+
\varphi_\theta\!\left(\widehat{S}_C(u_j^-)\right).
\end{equation}

Subtracting Equation~\ref{eq:cg3} from
Equation~\ref{eq:cg2} yields the jump recursion
\begin{align}
\label{eq:cg4}
&
\varphi_\theta\!\left(\widehat{S}_E(u_j)\right)
-
\varphi_\theta\!\left(\widehat{S}_E(u_j^-)\right)
\nonumber\\
&\qquad=
\varphi_\theta\!\left(\frac{n_j-d_j}{N}\right)
-
\varphi_\theta\!\left(\frac{n_j}{N}\right).
\end{align}

Since $\widehat{S}_E(t)=1$ before the first observed event and
$\varphi_\theta(1)=0$, summing the jumps up to time $t$ gives
\begin{align}
\label{eq:cg5}
\varphi_\theta\!\left(\widehat{S}_E(t)\right)
&=
\sum_{u_j\leq t}
\left[
\varphi_\theta\!\left(\widehat{S}_E(u_j)\right)
-
\varphi_\theta\!\left(\widehat{S}_E(u_j^-)\right)
\right]
\\
&=
\sum_{u_j\leq t}
\left[
\varphi_\theta\!\left(\frac{n_j-d_j}{N}\right)
-
\varphi_\theta\!\left(\frac{n_j}{N}\right)
\right].
\end{align}

Applying $\varphi_\theta^{-1}$ therefore gives the CG estimator
\begin{equation}
\label{eq:cg6}
\widehat{S}_E(t)
=
\varphi_\theta^{-1}
\left(
\sum_{u_j\leq t}
\left[
\varphi_\theta\!\left(\frac{n_j-d_j}{N}\right)
-
\varphi_\theta\!\left(\frac{n_j}{N}\right)
\right]
\right),
\qquad
0\leq t\leq t_{\max}.
\end{equation}
The empty sum is zero, and the estimator is undefined for
$t>t_{\max}$.

For the independence generator
$\varphi_{\mathrm{ind}}(u)=-\log u$, Equation~\ref{eq:cg6} becomes
\begin{align}
\widehat{S}_E(t)
&=
\exp\left(
-\sum_{u_j\leq t}
\left[
-\log\!\left(\frac{n_j-d_j}{N}\right)
+
\log\!\left(\frac{n_j}{N}\right)
\right]
\right)
\\
&=
\prod_{u_j\leq t}
\frac{n_j-d_j}{n_j}
=
\prod_{u_j\leq t}
\left(
1-\frac{d_j}{n_j}
\right),
\end{align}
which is the Kaplan-Meier estimator.

For the Clayton generator
$\varphi_\theta(u)=u^{-\theta}-1$~\citep{clayton1978model}, the CG
estimator becomes
\begin{equation}
\label{eq:cg7}
\widehat{S}_E(t)
=
\left[
1+
\sum_{u_j\leq t}
\left\{
\left(\frac{n_j-d_j}{N}\right)^{-\theta}
-
\left(\frac{n_j}{N}\right)^{-\theta}
\right\}
\right]^{-1/\theta}.
\end{equation}

\subsection{Theoretical analysis}
\label{app:margin_time_theory}

\citet{Rivest2001} study the CG estimator for $\widehat S_E(t)$ under dependent censoring. Without assuming the fitted Archimedean copula is the true copula, their estimator targets a marginal survival function $S_E^*(t)$ defined via their equation (8) (at page 6). Moreover, when the dependence is truly Archimedean with the same generator used to construct the estimator, $\widehat{S}_E(t)$ is uniformly consistent for $S_E(t)$~\citep[Theorem~1]{Rivest2001}. 

In this section, we provide theoretical properties of this margin time. We start the analysis by formally (re-)stating the calculation.

For any $c\ge 0$ with $S_E(c)>0$ and $\int_c^\infty S_E(u)\,du<\infty$, define
\begin{equation}
\label{eq:margin_time_full_def}
e^{\mathrm{margin}}(c;S_E)
\ :=\
c+\frac{\displaystyle \int_c^\infty S_E(u)\,du}{S_E(c)}.
\end{equation}
We estimate it by the plug-in CG estimator
\begin{equation}
\label{eq:margin_time_full_est}
\widehat e^{\mathrm{margin}}(c)
\ :=\ 
e^{\mathrm{margin}}(c;\widehat S_E)
\ = \
c+\frac{\displaystyle \int_c^\infty \widehat S_E(u)\,du}{\widehat S_E(c)}.
\end{equation}
Similarly, define the large-sample target
\begin{equation}
\label{eq:margin_time_full_target}
e^{*,\mathrm{margin}}(c)
\ :=\ 
e^{\mathrm{margin}}(c;S_E^*)
\ = \
c+\frac{\displaystyle \int_c^\infty S_E^*(u)\,du}{S_E^*(c)}.
\end{equation}
When the copula model is correctly specified, $S_E^*=S_E$ and hence
$e^{*,\mathrm{margin}}(c)=e^{\mathrm{margin}}(c;S_E)$.

To connect \eqref{eq:margin_time_full_def}--\eqref{eq:margin_time_full_est} to \citet{Rivest2001} results (proved on $[0,t_0)$),
without loss of generality,
we assume the right tail beyond some finite horizon (\ie~$t_0$) does not contribute.
Under such a setting, we can rewrite the integrals in
\eqref{eq:margin_time_full_est}-\eqref{eq:margin_time_full_target} using
\begin{align*}
    \int_c^\infty \widehat S_E(u)\,du \ & = \ \int_c^{t_0}\widehat S_E(u)\,du\\
    \int_c^\infty S_E^*(u)\,du \ & = \ \int_c^{t_0}S_E^*(u)\,du,
\end{align*}
so the theory reduces to controlling $\widehat S_E$ on the finite interval $[0,t_0)$
without any tail extrapolation.

\subsubsection{Margin time as conditional expectation}
\label{app:conditional_exp}

Now we prove our method in \eqref{eq:margin_time_full_def} corresponds to the conditional expectation.
This following theorem and proof are adapted from \citet[Theorem~B.1]{haider_effective_2020}. Here, we rephrase the proof for the completeness.

\begin{lemma}
\label{lem:cond_mean_survival_full}
Let $E\geq 0$ have survival function $S_E(t) = \Pr(E>t)$.
Fix $c\geq 0$ with $S_E(c)>0$ and assume $\int_c^\infty S_E(u)\,du<\infty$.
Then
\begin{align}
\label{eq:cond_mean_survival_full}
\mathbb{E}[E\mid E>c] \ = \ c+\frac{\int_c^\infty S_E(u)\,du}{S_E(c)}.
\end{align}
\end{lemma}

\begin{proof}
Define $(E-c)_+:=\max\{E-c,0\}$. For any nonnegative $Y$,
$\mathbb{E}[Y]=\int_0^\infty \Pr(Y>v)\,dv$. Apply this conditionally on $\{E>c\}$:
\begin{align*}
\mathbb{E}[(E-c)_+\mid E>c] \
&= \ \int_0^\infty \Pr((E-c)_+>v \mid E>c)\,dv \\
&= \ \int_0^\infty \Pr(E>c+v \mid E>c)\,dv \\
&= \ \int_0^\infty \frac{\Pr(E>c+v)}{\Pr(E>c)}\,dv \\
&= \ \frac{1}{S_E(c)}\int_0^\infty S_E(c+v)\,dv \\
&= \ \frac{1}{S_E(c)}\int_c^\infty S_E(u)\,du,
\end{align*}
where we substituted $u=c+v$. Finally, on $\{E>c\}$, $E=c+(E-c)$, hence
\begin{align*}
    \mathbb{E}[E\mid E>c] 
    &= \ c+\mathbb{E}[E-c\mid E>c] \\
    &= \ c+\mathbb{E}[(E-c)_+\mid E>c] \\
    &=\ c+\frac{\int_c^\infty S_E(u)\,du}{S_E(c)}.
\end{align*}
\end{proof}

\subsubsection{Consistency of the margin time}
\label{app:consistency}

To simplify the notation, we denote the numerator term in \eqref{eq:margin_time_full_def} as
\begin{align*}
    A(c;S) \ := \ \int_c^\infty S(u)\,du.
\end{align*}

\begin{lemma}
\label{thm:deterministic_bound_full}
Let $s_{\min}$ be a small positive constant that satisfies
$S(c)\ge s_{\min}>0$, $A(c;S)<\infty$, and $A(c;\widehat S)<\infty$.
Let
\[
\Delta A := A(c;\widehat S)-A(c;S)=\int_c^\infty\big(\widehat S(u)-S(u)\big)\,du,
\qquad
\Delta S(c):=\widehat S(c)-S(c).
\]
Assume $|\Delta S(c)|\le s_{\min}/2$ (so that $\widehat S(c)\ge s_{\min}/2$).
Then
\begin{equation}
\label{eq:det_bound_full}
\big|e^{\mathrm{margin}}(c;\widehat S)-e^{\mathrm{margin}}(c;S)\big|
\le
\frac{1}{s_{\min}}\,|\Delta A|
+
\frac{2A(c;S)}{s_{\min}^2}\,|\Delta S(c)|
+
\frac{2}{s_{\min}^2}\,|\Delta A|\,|\Delta S(c)|.
\end{equation}
In particular, for small perturbations, the last term is second order in the errors.
\end{lemma}

\begin{proof}
By definition,
\begin{align*}
e^{\mathrm{margin}}(c;\widehat S)-e^{\mathrm{margin}}(c;S) \ = \ \frac{A(c;\widehat S)}{\widehat S(c)}-\frac{A(c;S)}{S(c)}.
\end{align*}
Write $A(c;\widehat S)=A(c;S)+\Delta A$ and $\widehat S(c)=S(c)+\Delta S(c)$:
\begin{align*}
\frac{A(c;\widehat S)}{\widehat S(c)}-\frac{A(c;S)}{S(c)} \ = \ \frac{A(c;S)+\Delta A}{S(c)+\Delta S(c)}-\frac{A(c;S)}{S(c)}.
\end{align*}
Subtract and re-add $\frac{A(c;S)+\Delta A}{S(c)}$:
\begin{align}
\label{eq:det_full_step1}
\frac{A(c;S)+\Delta A}{S(c)+\Delta S(c)}-\frac{A(c;S)}{S(c)} \ 
&= \
\underbrace{\left(\frac{A(c;S)+\Delta A}{S(c)+\Delta S(c)}-\frac{A(c;S)+\Delta A}{S(c)}\right)}_{I}
+
\underbrace{\left(\frac{A(c;S)+\Delta A}{S(c)}-\frac{A(c;S)}{S(c)}\right)}_{II}.
\end{align}
Term II equals $\Delta A/S(c)$, hence $|(II)|\le |\Delta A|/s_{\min}$.

For term I,
\begin{align*}
(I)=(A(c;S)+\Delta A)\left(\frac{1}{S(c)+\Delta S(c)}-\frac{1}{S(c)}\right)
=(A(c;S)+\Delta A)\frac{-\Delta S(c)}{S(c)\,(S(c)+\Delta S(c))}.
\end{align*}
Taking absolute values and using $S(c)\ge s_{\min}$ and $S(c)+\Delta S(c)\ge s_{\min}/2$ gives
\begin{align*}
|(I)|
\le
|A(c;S)+\Delta A|\cdot \frac{2|\Delta S(c)|}{s_{\min}^2}
\le
\frac{2A(c;S)}{s_{\min}^2}|\Delta S(c)|+\frac{2}{s_{\min}^2}|\Delta A|\,|\Delta S(c)|.
\end{align*}
Combining the bounds for both terms yields \eqref{eq:det_bound_full}.
\end{proof}

\begin{corollary}[Consistency]
\label{cor:consistency_full}
Assume:
\begin{enumerate}
\item[(a)] Theorem~1 in \citet{Rivest2001} holds at some $t_0>c$, \ie $\widehat S_E(t)$ is uniformly
consistent for $S_E^*(t)$ on $[0,t_0)$;
\item[(b)] no tail contribution beyond $t_0$;
\item[(c)] $S_E^*(c)\ge s_{\min}>0$.
\end{enumerate}
Then $\widehat e^{\mathrm{margin}}(c)\xrightarrow{p} e^{*,\mathrm{margin}}(c)$.
If the copula model is correctly specified so that $S_E^*=S_E$,
then $\widehat e^{\mathrm{margin}}(c)\xrightarrow{p} e^{\mathrm{margin}}(c;S_E)$.
\end{corollary}

\begin{proof}
Because (b) assumes no tail contribution beyond $t_0$, we have
\begin{align*}
    \Delta A \ = \ \int_c^\infty(\widehat S_E(u)-S_E^*(u))\,du \ = \ \int_c^{t_0}(\widehat S_E(u)-S_E^*(u))\,du.
\end{align*}
Hence
\begin{align*}
    |\Delta A|
    & \le (t_0-c) \\
    \sup_{u\in[c,t_0]}|\widehat S_E(u)-S_E^*(u)|
    & \le  (t_0-c) \\
    \sup_{u\in[0,t_0]}|\widehat S_E(u)-S_E^*(u)|
    & \xrightarrow{p}0,
\end{align*}
by uniform consistency on $[0,t_0)$.
Also $|\Delta S(c)|=|\widehat S_E(c)-S_E^*(c)|\le \sup_{u\in[0,t_0]}|\widehat S_E(u)-S_E^*(u)|
\xrightarrow{p}0$.

Since $\sup_{u\in [0, t_0)} \left|\widehat{S}_E(u) - S_E^*(u) \right| \xrightarrow{p}0$ and $s_{\min} >0$, we have $\Pr\left(|\Delta S(c)| < \frac{s_{\min}}{2} \right) \xrightarrow{p}1$, hence $\widehat{S}_E(c) \geq \frac{s_{\min}}{2}$ with probability close to 1.

Then we can apply Lemma~\ref{thm:deterministic_bound_full} (because we have $|\Delta S(c)|\le s_{\min}/2$) with $S=S_E^*$ and $\widehat S=\widehat S_E$
to conclude $\widehat e^{\mathrm{margin}}(c)-e^{*,\mathrm{margin}}(c)\xrightarrow{p}0$.
The correctly specified case follows from $S_E^*=S_E$.
\end{proof}

\subsubsection{First-order expansion and delta method}
\label{app:delta_method_full}

In the preceding section, we prove the "law of large numbers property" of the margin time -- in the limit, we have asymptotic consistency. Now, we want to demonstrate the convergence rate.

Let us first define some notation for convenience.
Fix $c\ge 0$. Define
\begin{align*}
    \mathcal{T}_c(S) \ := \ e^{\mathrm{margin}}(c;S)
    \  = \ c+\frac{A(c;S)}{S(c)}.
\end{align*}

\begin{lemma}
\label{lem:frechet_expansion_full}
Let $S$ be a survival function with $S(c)>0$ and $A(c;S)<\infty$.
Let $h$ be any bounded function on $[c,\infty)$ such that $S(c)+h(c)>0$ and
$A(c;S+h)<\infty$.
Write $S_h:=S+h$ and define $\Delta A:=A(c;S_h)-A(c;S)=\int_c^\infty h(u)\,du$.
Then
\begin{equation}
\label{eq:exact_expansion_full}
\mathcal{T}_c(S_h)-\mathcal{T}_c(S) \ = \
\underbrace{\frac{1}{S(c)}\int_c^\infty h(u)\,du
-
\frac{A(c;S)}{S(c)^2}\,h(c)}_{\dot{\mathcal{T}}_{c;S}(h)}
+
R_{c;S}(h),
\end{equation}
where the remainder is
\begin{equation}
\label{eq:remainder_exact_full}
R_{c;S}(h) \ = \  -\frac{\Delta A\;h(c)}{S(c)^2}
+
\frac{A(c;S_h)\,h(c)^2}{S(c)^2\,(S(c)+h(c))}.
\end{equation}
In particular, if $|h(c)|\le S(c)/2$, then $R_{c;S}(h) \; = \; O\!\left(|h(c)|^2+|\Delta A|\,|h(c)|\right)$.
\end{lemma}

\begin{proof}
The proof is identical to the finite-horizon case, replacing $\int_c^{t_0}$ by $\int_c^\infty$.
Start from
\begin{align*}
    \mathcal{T}_c(S_h)-\mathcal{T}_c(S) \ =\ \frac{A(c;S_h)}{S_h(c)}-\frac{A(c;S)}{S(c)}.
\end{align*}
Add and subtract $A(c;S_h)/S(c)$ to obtain
\begin{align*}
    \frac{A(c;S_h)}{S_h(c)}-\frac{A(c;S)}{S(c)} \ 
    &= \ A(c;S_h)\left(\frac{1}{S_h(c)}-\frac{1}{S(c)}\right) + \frac{A(c;S_h)-A(c;S)}{S(c)} \\
    &= \ A(c;S_h)\left(\frac{1}{S_h(c)}-\frac{1}{S(c)}\right) + \frac{\Delta A}{S(c)}.
\end{align*}
Use the identity $\frac{1}{a+b}-\frac{1}{a} \; =\; -\frac{b}{a^2}+\frac{b^2}{a^2(a+b)}$ with
$a=S(c)$ and $b=h(c)$, and then decompose $A(c;S_h)=A(c;S)+\Delta A$.
Collecting terms yields \eqref{eq:exact_expansion_full}--\eqref{eq:remainder_exact_full}.
\end{proof}

Theorem~2 in \citet{Rivest2001} states that, for any $t_0$ satisfying
$\pi(t_0)>0$ and suitable bounded-derivative conditions,
$\sqrt{n}\bigl(\widehat S_E(t)-S_E^*(t)\bigr)$
converges weakly on $D[0,t_0)$ to a mean-zero Gaussian process, where
$\pi(t)=\Pr\!\left(\min(E,C)\ge t\right)$. Because the margin-time functional depends
only on $\widehat S_E$ over $[0,t_0]$, this functional CLT
is sufficient.

\begin{theorem}
\label{thm:delta_method_limit_full}
Assume:
\begin{enumerate}
\item[(a)] Rivest--Wells Theorem~2 holds at some $t_0>c$, \ie $\sqrt{n}(\widehat S_E-S_E^*)\Rightarrow \mathbb{G}_E$
on $D[0,t_0)$, where $\mathbb{G}_E$ has almost surely continuous sample paths;
\item[(b)] no tail contribution beyond $t_0$;
\item[(c)] $S_E^*(c)>0$.
\end{enumerate}
Then
\begin{equation}
\label{eq:delta_method_limit_full}
\sqrt{n}\Big(\widehat e^{\mathrm{margin}}(c)-e^{*,\mathrm{margin}}(c)\Big)
\quad \Rightarrow \quad
\dot{\mathcal{T}}_{c;S_E^*}(\mathbb{G}_E) \ = \ 
\frac{1}{S_E^*(c)}\int_c^\infty \mathbb{G}_E(u)\,du - \frac{\int_c^\infty S_E^*(u)\,du}{(S_E^*(c))^2}\,\mathbb{G}_E(c).
\end{equation}
Moreover, under no tail contribution assumption, the integrals in \eqref{eq:delta_method_limit_full}
equal $\int_c^{t_0}(\cdot)\,du$.
\end{theorem}

\begin{proof}
Let $h_n:=\widehat S_E-S_E^*$.
By Lemma~\ref{lem:frechet_expansion_full} applied with $S=S_E^*$ and $h=h_n$,
\begin{align*}
    \widehat e^{\mathrm{margin}}(c)-e^{*,\mathrm{margin}}(c) \ = \ \dot{\mathcal{T}}_{c;S_E^*}(h_n) + R_{c;S_E^*}(h_n).
\end{align*}
Multiply by $\sqrt{n}$:
\begin{align*}
    \sqrt{n}\Big(\widehat e^{\mathrm{margin}}(c)-e^{*,\mathrm{margin}}(c)\Big) \ =\ 
    \dot{\mathcal{T}}_{c;S_E^*}\big(\sqrt{n}h_n\big) + \sqrt{n}\,R_{c;S_E^*}(h_n).
\end{align*}

Under no tail contribution, $\widehat S_E(u)=S_E^*(u)=0$ for $u\ge t_0$ (w.p.\ $\to 1$),
so $\int_c^\infty h_n(u)\,du=\int_c^{t_0} h_n(u)\,du$ and likewise for $A(c;S_E^*)$.
Thus $\dot{\mathcal{T}}_{c;S_E^*}$ is linear and is continuous at every path that is continuous at $c$.

By Theorem~1 in \citet{Rivest2001}, $\widehat S_E$ is uniformly consistent for $S_E^*$ on $[0,t_0)$,
so $\|h_n\|_{\infty,[0,t_0]} \xrightarrow{p}0$.
The remainder in Lemma~\ref{lem:frechet_expansion_full} is second order in $h_n(c)$ and $\Delta A=\int_c^{t_0}h_n$,
hence $\sqrt{n}R_{c;S_E^*}(h_n)=o_p(1)$ under the same tightness argument implied by the functional CLT in (a).

Since $\sqrt{n}h_n\Rightarrow \mathbb{G}_E$ on $D[0,t_0)$ and
$\mathbb{G}_E$ has almost surely continuous sample paths,
$\dot{\mathcal{T}}_{c;S_E^*}$ is continuous at $\mathbb{G}_E$ almost surely.
The continuous mapping theorem therefore yields
\[
\dot{\mathcal{T}}_{c;S_E^*}\bigl(\sqrt{n}h_n\bigr)
\Rightarrow
\dot{\mathcal{T}}_{c;S_E^*}(\mathbb{G}_E).
\]
Combining this with
$\sqrt{n}R_{c;S_E^*}(h_n)=o_p(1)$
and Slutsky's theorem gives
\eqref{eq:delta_method_limit_full}.
\end{proof}

\section{Experimental details}\label{app:experimental_details}

\subsection{Synthetic experiment details}
\label{app:synthetic_experiments}

We evaluate the behavior of survival evaluation metrics under controlled censoring mechanisms using a fully synthetic data-generating process (DGP) with known ground-truth event times. The primary goal of these experiments is to isolate the effect of dependent censoring on metric bias while keeping the predictive model fixed.

All experiments use Weibull marginal distributions for both event and censoring times and induce dependence through a copula function. Synthetic datasets are generated using a linear Weibull DGP, and all experimental factors other than the evaluation metric are held fixed.

\textbf{Covariates.}
For each experiment, we generate $N$ independent covariate vectors
\[
\bm{x}_i \sim \mathrm{Unif}([0,1]^d),
\]
with $d=10$ and $N=10{,}000$. Covariates are fixed across different censoring dependence levels for each random seed to isolate the effect of the censoring mechanism.

\textbf{Event and censoring models.}
Both the event time $E$ and censoring time $C$ follow Weibull distributions with covariate-dependent hazards. Conditional on $\bm{x}_i$, the hazard and survival functions take the form
\begin{align}
h(t \mid \bm{x}_i) &= \left(\frac{v}{\rho}\right)
\left(\frac{t}{\rho}\right)^{v-1}
\exp\big(g(\bm{x}_i)\big), \\
S(t \mid \bm{x}_i) &= \exp\!\left(
- \left(\frac{t}{\rho}\right)^{v}
\exp\big(g(\bm{x}_i)\big)
\right).
\end{align}

We use the following parameters throughout:
\[
(v_E, \rho_E) = (4, 17), \qquad
(v_C, \rho_C) = (3, 12).
\]

The risk functions are linear:
\[
g_E(\bm{x}) = \bm{x}^\top \bm{\beta}_E,
\qquad
g_C(\bm{x}) = \bm{x}^\top \bm{\beta}_C,
\]
where $\bm{\beta}_E, \bm{\beta}_C \in \mathbb{R}^d$ are sampled independently for each random seed from $\mathrm{Unif}([-1,1]^d)$.

\textbf{Copula-based dependence.}
Dependence between event and censoring times is introduced using copula-based constructions. For each instance $i$, a pair of uniform random variables
\[
(u_i, v_i) \in [0,1]^2
\]
is sampled from a copula with a specified Kendall’s $\tau$.

For Archimedean copulas (Clayton and Frank), $(u_i, v_i)$ is sampled directly using the standard Archimedean construction, with the copula parameter chosen to match the target Kendall’s $\tau$ via the corresponding analytical relationship. For the Gaussian copula, dependence is induced by first sampling a bivariate normal vector $(z_{1i},z_{2i})$ with zero mean, unit variance, and correlation coefficient $\rho$. The correlation $\rho$ is chosen such that the resulting Kendall’s $\tau$ matches the target value. The uniform variables are then obtained via the probability integral transform:
\[
u_i=\Phi(z_{1i}), \qquad v_i=\Phi(z_{2i}).
\]
where $\Phi(\cdot)$ denotes the standard normal cumulative distribution function.

When $\tau = 0$, $u_i$ and $v_i$ are sampled independently, yielding
conditional independence between $E$ and $C$ given $\bm{X}$.

Event and censoring times are subsequently generated via inverse transform sampling:
\[
e_i = S_E^{-1}(v_i \mid \bm{x}_i), \qquad
c_i = S_C^{-1}(u_i \mid \bm{x}_i).
\]

\textbf{Censoring rate calibration.}
To ensure that differences in metric behavior are not confounded by changes in censoring severity, the censoring distribution is calibrated to achieve approximately matched marginal censoring rates across copula families and dependence strengths.

This is accomplished by introducing a multiplicative scaling factor on the censoring hazard. For each copula family, a grid of scaling factors is evaluated in pilot simulations, and the resulting empirical censoring rates are recorded. Scaling factors are then selected to span a fixed set of target censoring-rate intervals, ensuring that each copula family contributes comparable censoring levels despite differences in dependence structure.

The selected scaling factors are reused across random seeds, yielding stable censoring rates with low variability while preserving the intended dependence between event and censoring times.

\textbf{Observed data.}
The observed time and event indicator are given by
\[
t_i = \min(e_i, c_i), \qquad
\delta_i = \mathbb{I}\{e_i \le c_i\}.
\]
For oracle evaluation, the true event times $e_i$ are retained for all instances.

\textbf{Model fitting.}
For each simulated dataset, we fit a CoxPH model using the observed data $(t_i, \delta_i, \bm{x}_i)$. Survival curves $\widehat S(t \mid \bm{x}_i)$ are obtained using the Breslow estimator. The CoxPH model specification and hyperparameters are held fixed across experiments, while the model is refitted for each simulated dataset.

\textbf{Metric evaluation.}
We compare two evaluation approaches:
\begin{itemize}
\item \textbf{IPCW Brier Score}, computed using inverse probability of censoring weights as implemented in the \texttt{SurvivalEVAL} package~\citep{qi_survivaleval_2024}.
\item \textbf{Dependent Brier Score}, computed using the proposed dependent Brier score based on CG margin-time imputation with uncertainty weighting and with the copula family and dependence strength assumed known.
\end{itemize}

Metric error is defined as the absolute deviation from the oracle Brier score computed using the true event times. All results are averaged over 10 random seeds. Empirical censoring rates are additionally reported to verify calibration stability across dependence levels.

\subsection{Datasets and preprocessing}
\label{app:data_generation_and_processing}

\textbf{WHAS:} Worcester Heart Attack Study dataset (WHAS)~\citep{hosmer2008applied} contains 500 patients with acute myocardial infarction, focusing on the time to death post-hospital admission. The data was already post-processed and can be downloaded from the \texttt{scikit-survival} package~\citep{polsterl2020scikit}.

\textbf{METABRIC:} The Molecular Taxonomy of Breast Cancer International Consortium (METABRIC) dataset \citep{Curtis2012} contains survival information for breast cancer patients. This dataset includes a diverse range of feature sets that encompass clinical traits, expression profiles, copy number variation (CNV) profiles, and single-nucleotide polymorphism (SNP) genotypes. All these features are derived from breast tumor samples collected during the METABRIC trial. The dataset can be downloaded from \url{https://www.cbioportal.org/study/summary?id=brca_metabric}.

\textbf{Churn:} Customer churn prediction dataset (Churn) focuses on predicting customer attrition. Each record in the file corresponds to a customer and contains the same set of features, such as age, gender, tenure, usage frequency, support calls, payment delay, subscription type, contract length, total spend, and last interaction. We apply one-hot encoding on the US region feature and exclude subjects who are censored at time 0. The dataset can be downloaded from the PySurvival package~\citep{fotso2019pysurvival}.

\textbf{GBSG:} The German Breast Cancer Study Group (GBSG) dataset~\citep{royston2013external} includes 686 patients with node-positive breast cancer and complete prognostic information. It is publicly available in R's \texttt{survival} package~\citep{therneau2024survival}. While the original GBSG dataset features a higher censoring rate and a broader set of variables, we use a modified version that combines the dataset with uncensored cases from the Rotterdam dataset. This results in fewer variables, a lower censoring rate, and a larger sample size than the original. This combined dataset, which was already post-processed in the DeepSurv study~\citep{katzman_deepsurv_2018}, is available for download at \url{https://github.com/jaredleekatzman}.

\textbf{NACD:} The Northern Alberta Cancer Dataset (NACD)~\citep{haider_effective_2020} contains 2402 patients and 53 features. It is a conglomerate of many different cancer patients, including lung, colorectal, head and neck, esophageal, stomach, and other cancers. The event of interest in this dataset is failure time. We drop patients with negative or zero survival time leaving 2,396 patients. There are no missing values in this dataset. The dataset can be downloaded from \url{http://pssp.srv.ualberta.ca} under "Public Predictors".

\textbf{FLCHAIN:} The Serum Free Light Chain (FLCHAIN) dataset is a stratified random sample comprising half of the participants from a study investigating the association between serum free light chain (FLC) levels and mortality~\citep{dispenzieri2012use}. It is available in R’s \texttt{survival} package~\citep{therneau2024survival}. After downloading, we perform several preprocessing steps: we remove three subjects with events occurring at time zero, impute missing values in the \texttt{creatinine} feature using its median, and exclude the \texttt{chapter} feature—which encodes the cause of death according to ICD chapter headings—since it is only defined for deceased (uncensored) subjects and would therefore leak the event indicator to the model.

\textbf{SUPPORT:} The Study to Understand Prognoses, Preferences, Outcomes, and Risks of Treatment (SUPPORT) dataset~\citep{knaus_support_1995} includes 9,105 participants and was designed to investigate survival outcomes and clinical decision-making among seriously ill hospitalized patients. The dataset contains a substantial proportion of missing values across many features. We obtained the dataset from \citet{qi2024conformalized} and follow their preprocessing steps.

\textbf{Employee:} Employee dataset contains employee activity information that can be used to predict when an employee will quit. The dataset can be downloaded from the PySurvival package~\citep{fotso2019pysurvival}. It contains duplicate entries; after dropping these duplicates, the number of subjects in the dataset is reduced from 14,999 to 11,991. We also apply one-hot encoding to the department information.

\textbf{MIMIC-IV:} The Medical Information Mart for Intensive Care IV (MIMIC-IV)~\citep{johnson_mimic_2023} contains critical care data from patients admitted to hospitals and intensive care units (ICU). We create a version using the MIMIC-IV database, which contains patients who are alive at least 24 hours after being admitted to ICU. Their date of death is derived from hospital records or state records, which means the cause of mortality is not limited to the reason for ICU admission. We follow the instructions in \citep{qi_effective_2023} to process the dataset, and the code is available in their GitHub repository: \url{https://github.com/shi-ang/CensoredMAE}.

\textbf{SEER:} The SEER Program dataset~\citep{ries_cancer_2003} is a comprehensive collection of cancer patient data from approximately 49\% of the U.S. population. Our study focuses on three subsets: SEER-brain, SEER-liver, and SEER-stomach, containing data on patients diagnosed with brain, liver, and stomach cancers, respectively. The goal is to model the time from diagnosis to failure events, such as death or disease progression. We preprocess the dataset as in~\citep{qi2024conformalized}. Features analyzed include age, sex, cancer stage, grade, and various clinical factors. We excluded features with over 70\% missing data, patients without follow-up times, duplicates, and those with a survival time of zero. The SEER cohort is available for download at \url{www.seer.cancer.gov}.

\begin{figure}[!ht]
\centering
\includegraphics[width=0.9\linewidth]{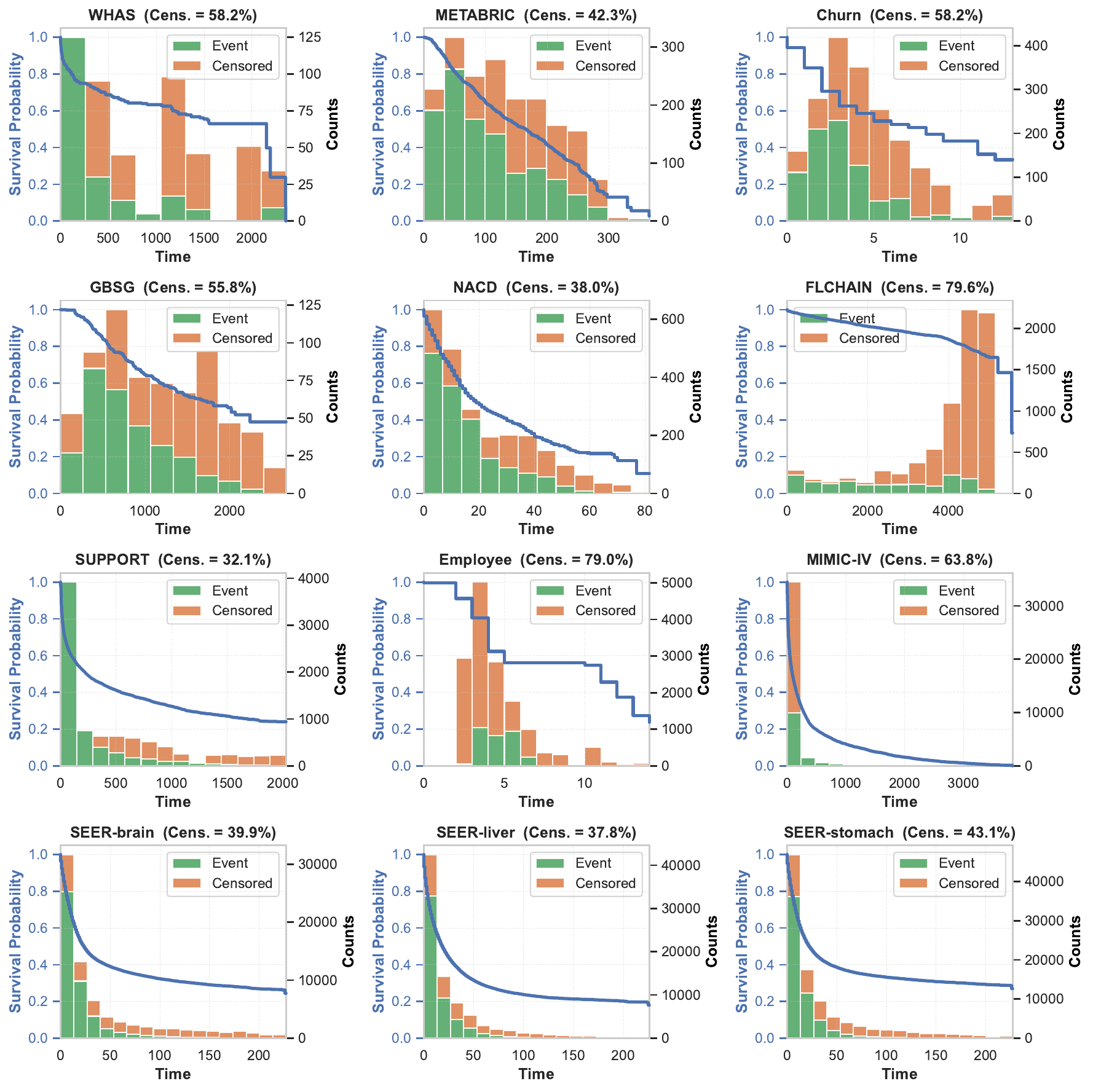}
\caption{Empirical KM survival curve and event/censoring time distribution for the generated semi-synthetic datasets used in training and evaluation.}
\label{fig:semi_synth_distribution}
\end{figure}

Figure~\ref{fig:semi_synth_distribution} shows the empirical KM survival curves and the distribution of event and censoring times for the generated semi-synthetic datasets used for training and evaluation. Reported censoring rates are computed from the observed indicators.

\subsection{Survival learners}
\label{app:survival_learners}

\textbf{CoxPH:} The Cox Proportional Hazards model (CoxPH) is a semi-parametric survival analysis model \citep{cox_regression_1972}, consisting of a non-parametric baseline hazard function $h_0(t)$ and a parametric partial hazard function $f(\bm{\theta}, \bx_i)$, typically estimated using the Breslow method~\citep{breslow_analysis_1975}. The hazard function is $h(t\mid\bx_i) = h_0(t) \exp(f(\bm{\theta}, \bx_i))$, where $f$ is linear. The MLE of $\widehat{\bm{\theta}}$ maximizes the Cox partial log-likelihood \citep{cox_regression_1972}. CoxPH is implemented from the \texttt{scikit-survival} package~\citep{polsterl2020scikit}.

\textbf{GBSA:} GBSA \citep{Ridgeway1999} is a likelihood-based boosting model that optimizes an $\ell_2$-norm penalized Cox partial log-likelihood \citep{cox_regression_1972}. It updates candidate variables flexibly at each boosting step using an offset-based gradient boosting method, unlike traditional gradient boosting, which updates either a single component or all features for fitting the gradient. GBSA is implemented from the \texttt{scikit-survival} package~\citep{polsterl2020scikit}.

\textbf{RSF:} Random Survival Forests (RSF)~\citep{ishwaran_random_2008} extend decision trees for survival analysis by constructing multiple trees on bootstrapped samples. Each tree predicts an individual's survival function, with splits based on criteria like the log-rank test. RSF handles censored data as right-censored and accounts for it during tree construction, with final predictions as the average of all trees' survival functions. RSF is implemented from the \texttt{scikit-survival} package~\citep{polsterl2020scikit}.

\textbf{DeepSurv:} DeepSurv~\citep{katzman_deepsurv_2018} is a neural network implementation of the CoxPH model of the form $h\br{t \mid \bx_i} = h_0\br{t} \exp \br{f\br{\bth,\bx_i}}$, where $f\br{\bth,\bx_i}$ denotes a risk score as a nonlinear function of the features, \ie $f\br{\bth,\bx_i} = \sigma(\bx_i\bth)$, where $\sigma$ is a nonlinear activation function. Popular functions include the hyperbolic tangent (Tanh) function or the Rectified Linear Unit (ReLU). The MLE for $\widehat{\bth}$ is derived by numerically maximizing the partial Cox log-likelihood. DeepSurv is implemented from \url{https://github.com/shi-ang/BNN-ISD/}.

\textbf{MTLR:} Multi-Task Logistic Regression (MTLR)~\citep{NIPS2011_1019c809} is a neural network implementation of a discrete-time model that predicts the survival distribution using a sequence of dependent logistic regressors. The number of discrete times is determined by the square root of the number of uncensored patients, and we use quantiles to divide those uncensored instances evenly into each time interval. MTLR is implemented from \url{https://github.com/shi-ang/BNN-ISD/}.

Table \ref{tab:baseline_hyperparameters} reports the selected hyperparameters for the survival learners.

\begin{table}[!ht]
\centering
\caption{Hyperparameters for the survival learners.}
\label{tab:baseline_hyperparameters}
\resizebox{0.3\textwidth}{!}{%
\begin{tabular}{lll}
\toprule
{Model} & {Parameter} & {Value} \\ 
\midrule
{CoxPH} & alpha & 0.01 \\
 & ties & 'Breslow' \\
 & n\_iter & 100 \\
 & tol & 1e-9 \\ \midrule

{GBSA} & n\_estimators & 100 \\
 & max\_depth & 1 \\
 & min\_samples\_split & 2 \\
 & min\_samples\_leaf & 1 \\
 & max\_features & 'sqrt' \\
 & sub\_sample & 0.8 \\
 & random\_state & 0 \\ \midrule

{RSF} & n\_estimators & 100 \\
 & max\_depth & 1 \\
 & min\_samples\_split & 2 \\
 & min\_samples\_leaf & 1 \\
 & max\_features & 'sqrt' \\
 & random\_state & 0 \\ \midrule

{DeepSurv} & hidden\_size & 100 \\
 & verbose & False \\
 & lr & 0.001 \\
 & c1 & 0.01 \\
 & num\_epochs & 1000 \\
 & batch\_size & 32 \\
 & dropout & 0.25 \\
 & early\_stop & True \\
 & patience & 10 \\ \midrule

{MTLR} 
 & verbose & False \\
 & lr & 0.001 \\
 & c1 & 0.01 \\
 & num\_epochs & 1000 \\
 & batch\_size & 32 \\
 & early\_stop & True \\
 & patience & 10 \\

\bottomrule
\end{tabular}%
}
\end{table}

\subsection{Implementation details}
\label{app:implementation_details}

\textbf{Subsampling.} For computational efficiency, we subsample the largest semi-synthetic datasets while preserving their event-time distributions. For MIMIC-IV, SEER, and Employee, we draw $N=10{,}000$ rows stratified by event status and discretized time intervals.

\textbf{Reproducibility.} Experiments used Python 3.9, PyTorch 1.13.1, NumPy 1.24.3, and Pandas 1.5.3, with tensor computations in double precision (fp64). They were run on a workstation with an AMD Ryzen 5 5600X 3.70 GHz CPU, 32 GB RAM, and an NVIDIA GeForce RTX 2060 (6 GB) GPU using CUDA 11.7. Parts of the pipeline and preprocessing utilities build on the \textit{MENSA} codebase~\citep{lillelund_mensa_2026}.

\textbf{Copula estimation details.} For semi-synthetic experiments, copula and marginal parameters were optimized using Adam~\citep{KingBa15} for up to $3\times10^{4}$ full-batch epochs. Learning rates were $10^{-3}$ for the Weibull marginals and $10^{-2}$ for the copula parameter $\theta$. Early stopping was triggered when the validation NLL failed to improve for 100 epochs and $\theta$ changed negligibly across successive epochs.

\section{Additional results}\label{app:additional_results}

\subsection{Computational analysis}
\label{app:computational_analysis}

\textbf{Time complexity.} Table~\ref{tab:copula_fitting} reports the runtime of copula fitting, the dominant cost of dependent metric evaluation. For each candidate copula family, two Weibull marginals and a copula parameter are fitted jointly, followed by validation-based model selection. Runtime ranges from approximately 5 to 75 minutes. After fitting, the dependent metrics add less than 1ms relative to standard estimators across all datasets.

\textbf{Space complexity.} Table~\ref{tab:copula_fitting} also reports the increase in peak PyTorch-allocated GPU tensor memory during fitting. This excludes CPU RAM, non-PyTorch allocations, and reserved but unallocated GPU memory. The increase ranges from approximately 0.3 to 11MiB. Fitted models are discarded afterward and therefore do not affect metric-evaluation memory.

\begin{table}[!ht]
\centering
\caption{Copula fitting time and incremental peak PyTorch GPU tensor memory across datasets under the \emph{Original} feature selection strategy. Fitting includes training Clayton and Frank copula models jointly with their Weibull marginals.}
\label{tab:copula_fitting}
\resizebox{0.49\textwidth}{!}{
\begin{tabular}{lcc}
\toprule
Dataset & Copula Time (min) & Copula Memory (MiB) \\
\midrule
METABRIC        & 15.67 & 1.12 \\
GBSG            & 5.22  & 0.43 \\
NACD            & 14.76 & 1.91 \\
SUPPORT         & 67.36 & 5.90 \\
FLChain         & 55.98 & 5.40 \\
WHAS            & 10.61 & 0.35 \\
Employee        & 54.41 & 6.08 \\
Churn           & 5.59  & 1.32 \\
MIMIC-IV (all)  & 74.24 & 10.68 \\
SEER (brain)    & 28.97 & 5.72 \\
SEER (liver)    & 68.35 & 5.92 \\
SEER (stomach)  & 38.01 & 5.92 \\
\bottomrule
\end{tabular}
}
\end{table}

\subsection{Estimation error}
\label{app:estimation_error}

Table~\ref{tab:estimation_error_mean_pm_std} reports the full bias and variability results corresponding to the main text estimation error analysis.

\begin{table*}[!ht]
\centering
\caption{Mean ($\pm$ SD) absolute estimation error of IBS-IPCW and proposed dependent metrics across semi-synthetic datasets and feature selection strategies, averaged over 5 survival learners and 10 experiments.}
\label{tab:estimation_error_mean_pm_std}
\begin{subtable}{\textwidth}
\centering
\resizebox{0.9\textwidth}{!}{
\begin{tabular}{llcccccc}
& & WHAS & METABRIC & Churn & GBSG & NACD & FLCHAIN \\
\midrule
\multirow{5}{*}{Original} & IPCW (KM) & 0.073 $\pm$ 0.038 & 0.061 $\pm$ 0.012 & 0.117 $\pm$ 0.016 & 0.125 $\pm$ 0.039 & 0.092 $\pm$ 0.013 & 0.156 $\pm$ 0.008 \\
 & IPCW (CoxPH) & 0.106 $\pm$ 0.070 & 0.051 $\pm$ 0.012 & 0.115 $\pm$ 0.019 & 0.129 $\pm$ 0.040 & 0.148 $\pm$ 0.106 & 0.162 $\pm$ 0.008 \\
 & Dep (KM) & 0.055 $\pm$ 0.025 & 0.047 $\pm$ 0.011 & 0.120 $\pm$ 0.012 & 0.130 $\pm$ 0.038 & 0.091 $\pm$ 0.010 & 0.135 $\pm$ 0.008 \\
 & Dep (CG) & 0.055 $\pm$ 0.025 & 0.047 $\pm$ 0.011 & 0.120 $\pm$ 0.013 & 0.126 $\pm$ 0.037 & 0.091 $\pm$ 0.010 & 0.135 $\pm$ 0.008 \\
 & Dep (CG)$^{\text{UW}}$ & 0.063 $\pm$ 0.025 & 0.049 $\pm$ 0.010 & 0.124 $\pm$ 0.014 & 0.134 $\pm$ 0.037 & 0.082 $\pm$ 0.011 & 0.090 $\pm$ 0.008 \\
\midrule
\multirow{5}{*}{Top-5} & IPCW (KM) & 0.071 $\pm$ 0.037 & 0.062 $\pm$ 0.012 & 0.121 $\pm$ 0.018 & 0.124 $\pm$ 0.039 & 0.103 $\pm$ 0.014 & 0.133 $\pm$ 0.008 \\
 & IPCW (CoxPH) & 0.065 $\pm$ 0.037 & 0.061 $\pm$ 0.013 & 0.114 $\pm$ 0.016 & 0.126 $\pm$ 0.038 & 0.098 $\pm$ 0.014 & 0.138 $\pm$ 0.008 \\
 & Dep (KM) & 0.055 $\pm$ 0.024 & 0.048 $\pm$ 0.011 & 0.120 $\pm$ 0.013 & 0.130 $\pm$ 0.038 & 0.098 $\pm$ 0.010 & 0.108 $\pm$ 0.008 \\
 & Dep (CG) & 0.055 $\pm$ 0.024 & 0.048 $\pm$ 0.011 & 0.119 $\pm$ 0.014 & 0.126 $\pm$ 0.037 & 0.097 $\pm$ 0.012 & 0.108 $\pm$ 0.008 \\
 & Dep (CG)$^{\text{UW}}$ & 0.065 $\pm$ 0.024 & 0.049 $\pm$ 0.010 & 0.125 $\pm$ 0.016 & 0.134 $\pm$ 0.037 & 0.089 $\pm$ 0.012 & 0.066 $\pm$ 0.008 \\
\midrule
\multirow{5}{*}{Top-10} & IPCW (KM) & 0.071 $\pm$ 0.038 & 0.061 $\pm$ 0.011 & 0.119 $\pm$ 0.016 & 0.125 $\pm$ 0.039 & 0.098 $\pm$ 0.013 & 0.143 $\pm$ 0.013 \\
 & IPCW (CoxPH) & 0.082 $\pm$ 0.051 & 0.051 $\pm$ 0.012 & 0.111 $\pm$ 0.015 & 0.130 $\pm$ 0.041 & 0.093 $\pm$ 0.017 & 0.150 $\pm$ 0.014 \\
 & Dep (KM) & 0.054 $\pm$ 0.025 & 0.047 $\pm$ 0.011 & 0.121 $\pm$ 0.012 & 0.131 $\pm$ 0.038 & 0.095 $\pm$ 0.010 & 0.123 $\pm$ 0.017 \\
 & Dep (CG) & 0.054 $\pm$ 0.025 & 0.047 $\pm$ 0.011 & 0.121 $\pm$ 0.012 & 0.126 $\pm$ 0.037 & 0.095 $\pm$ 0.010 & 0.123 $\pm$ 0.017 \\
 & Dep (CG)$^{\text{UW}}$ & 0.063 $\pm$ 0.025 & 0.049 $\pm$ 0.011 & 0.125 $\pm$ 0.014 & 0.134 $\pm$ 0.037 & 0.086 $\pm$ 0.011 & 0.078 $\pm$ 0.009 \\
\midrule
\multirow{5}{*}{Rand.\ 25\%} & IPCW (KM) & 0.090 $\pm$ 0.030 & 0.061 $\pm$ 0.018 & 0.129 $\pm$ 0.014 & 0.153 $\pm$ 0.032 & 0.109 $\pm$ 0.013 & 0.140 $\pm$ 0.020 \\
 & IPCW (CoxPH) & 0.091 $\pm$ 0.031 & 0.056 $\pm$ 0.018 & 0.122 $\pm$ 0.013 & 0.157 $\pm$ 0.036 & 0.111 $\pm$ 0.017 & 0.147 $\pm$ 0.013 \\
 & Dep (KM) & 0.081 $\pm$ 0.022 & 0.049 $\pm$ 0.017 & 0.126 $\pm$ 0.009 & 0.153 $\pm$ 0.026 & 0.101 $\pm$ 0.010 & 0.140 $\pm$ 0.012 \\
 & Dep (CG) & 0.081 $\pm$ 0.022 & 0.049 $\pm$ 0.017 & 0.117 $\pm$ 0.009 & 0.147 $\pm$ 0.026 & 0.099 $\pm$ 0.010 & 0.140 $\pm$ 0.012 \\
 & Dep (CG)$^{\text{UW}}$ & 0.095 $\pm$ 0.026 & 0.050 $\pm$ 0.017 & 0.131 $\pm$ 0.009 & 0.161 $\pm$ 0.026 & 0.091 $\pm$ 0.010 & 0.032 $\pm$ 0.013 \\
\bottomrule
\end{tabular}}
\end{subtable}

\vspace{0.5cm}

\begin{subtable}{\textwidth}
\centering
\resizebox{0.9\textwidth}{!}{
\begin{tabular}{llcccccc}
& & SUPPORT & Employee & MIMIC (IV) & SEER (brain) & SEER (liver) & SEER (stomach) \\
\midrule
\multirow{5}{*}{Original} & IPCW (KM) & 0.108 $\pm$ 0.052 & 0.152 $\pm$ 0.009 & 0.026 $\pm$ 0.016 & 0.077 $\pm$ 0.005 & 0.104 $\pm$ 0.013 & 0.119 $\pm$ 0.018 \\
 & IPCW (CoxPH) & 0.109 $\pm$ 0.052 & 0.100 $\pm$ 0.006 & 0.039 $\pm$ 0.027 & 0.055 $\pm$ 0.005 & 0.097 $\pm$ 0.015 & 0.095 $\pm$ 0.012 \\
 & Dep (KM) & 0.111 $\pm$ 0.051 & 0.148 $\pm$ 0.007 & 0.023 $\pm$ 0.012 & 0.089 $\pm$ 0.005 & 0.118 $\pm$ 0.004 & 0.125 $\pm$ 0.008 \\
 & Dep (CG) & 0.109 $\pm$ 0.051 & 0.154 $\pm$ 0.007 & 0.024 $\pm$ 0.012 & 0.084 $\pm$ 0.006 & 0.106 $\pm$ 0.004 & 0.112 $\pm$ 0.008 \\
 & Dep (CG)$^{\text{UW}}$ & 0.093 $\pm$ 0.051 & 0.152 $\pm$ 0.008 & 0.024 $\pm$ 0.013 & 0.067 $\pm$ 0.005 & 0.092 $\pm$ 0.004 & 0.097 $\pm$ 0.008 \\
\midrule
\multirow{5}{*}{Top-5} & IPCW (KM) & 0.109 $\pm$ 0.052 & 0.155 $\pm$ 0.010 & 0.027 $\pm$ 0.017 & 0.076 $\pm$ 0.005 & 0.109 $\pm$ 0.014 & 0.124 $\pm$ 0.018 \\
 & IPCW (CoxPH) & 0.109 $\pm$ 0.052 & 0.103 $\pm$ 0.005 & 0.027 $\pm$ 0.018 & 0.083 $\pm$ 0.022 & 0.109 $\pm$ 0.017 & 0.128 $\pm$ 0.020 \\
 & Dep (KM) & 0.112 $\pm$ 0.051 & 0.149 $\pm$ 0.007 & 0.026 $\pm$ 0.011 & 0.090 $\pm$ 0.005 & 0.120 $\pm$ 0.005 & 0.126 $\pm$ 0.008 \\
 & Dep (CG) & 0.109 $\pm$ 0.051 & 0.156 $\pm$ 0.007 & 0.027 $\pm$ 0.013 & 0.085 $\pm$ 0.005 & 0.111 $\pm$ 0.005 & 0.116 $\pm$ 0.008 \\
 & Dep (CG)$^{\text{UW}}$ & 0.094 $\pm$ 0.051 & 0.154 $\pm$ 0.008 & 0.025 $\pm$ 0.013 & 0.068 $\pm$ 0.005 & 0.096 $\pm$ 0.005 & 0.101 $\pm$ 0.008 \\
\midrule
\multirow{5}{*}{Top-10} & IPCW (KM) & 0.109 $\pm$ 0.052 & 0.152 $\pm$ 0.009 & 0.026 $\pm$ 0.017 & 0.077 $\pm$ 0.005 & 0.103 $\pm$ 0.013 & 0.119 $\pm$ 0.017 \\
 & IPCW (CoxPH) & 0.109 $\pm$ 0.051 & 0.101 $\pm$ 0.005 & 0.031 $\pm$ 0.026 & 0.055 $\pm$ 0.005 & 0.097 $\pm$ 0.015 & 0.099 $\pm$ 0.015 \\
 & Dep (KM) & 0.111 $\pm$ 0.051 & 0.147 $\pm$ 0.007 & 0.023 $\pm$ 0.011 & 0.090 $\pm$ 0.005 & 0.118 $\pm$ 0.004 & 0.124 $\pm$ 0.008 \\
 & Dep (CG) & 0.109 $\pm$ 0.051 & 0.153 $\pm$ 0.007 & 0.024 $\pm$ 0.011 & 0.085 $\pm$ 0.006 & 0.107 $\pm$ 0.004 & 0.111 $\pm$ 0.008 \\
 & Dep (CG)$^{\text{UW}}$ & 0.093 $\pm$ 0.051 & 0.152 $\pm$ 0.008 & 0.025 $\pm$ 0.013 & 0.068 $\pm$ 0.005 & 0.093 $\pm$ 0.004 & 0.097 $\pm$ 0.009 \\
\midrule
\multirow{5}{*}{Rand.\ 25\%} & IPCW (KM) & 0.142 $\pm$ 0.058 & 0.161 $\pm$ 0.016 & 0.016 $\pm$ 0.009 & 0.111 $\pm$ 0.027 & 0.128 $\pm$ 0.015 & 0.132 $\pm$ 0.017 \\
 & IPCW (CoxPH) & 0.142 $\pm$ 0.058 & 0.110 $\pm$ 0.010 & 0.017 $\pm$ 0.011 & 0.111 $\pm$ 0.028 & 0.197 $\pm$ 0.105 & 0.133 $\pm$ 0.011 \\
 & Dep (KM) & 0.145 $\pm$ 0.057 & 0.137 $\pm$ 0.008 & 0.031 $\pm$ 0.013 & 0.113 $\pm$ 0.017 & 0.127 $\pm$ 0.004 & 0.135 $\pm$ 0.011 \\
 & Dep (CG) & 0.141 $\pm$ 0.057 & 0.120 $\pm$ 0.008 & 0.032 $\pm$ 0.014 & 0.102 $\pm$ 0.016 & 0.109 $\pm$ 0.005 & 0.120 $\pm$ 0.013 \\
 & Dep (CG)$^{\text{UW}}$ & 0.125 $\pm$ 0.057 & 0.147 $\pm$ 0.009 & 0.021 $\pm$ 0.007 & 0.089 $\pm$ 0.017 & 0.098 $\pm$ 0.004 & 0.106 $\pm$ 0.013 \\
\bottomrule
\end{tabular}}
\end{subtable}

\end{table*}

\end{document}